%% file: main.tex
\documentclass[10pt,twocolumn,letterpaper]{article}

\usepackage[pagenumbers]{cvpr}%

\input{preamble}

\definecolor{cvprblue}{rgb}{0.21,0.49,0.74}
\usepackage[pagebackref,breaklinks,colorlinks,allcolors=cvprblue]{hyperref}

\title{\texorpdfstring{\raisebox{-0.18em}[0pt][0pt]{\includegraphics[height=1.05em]{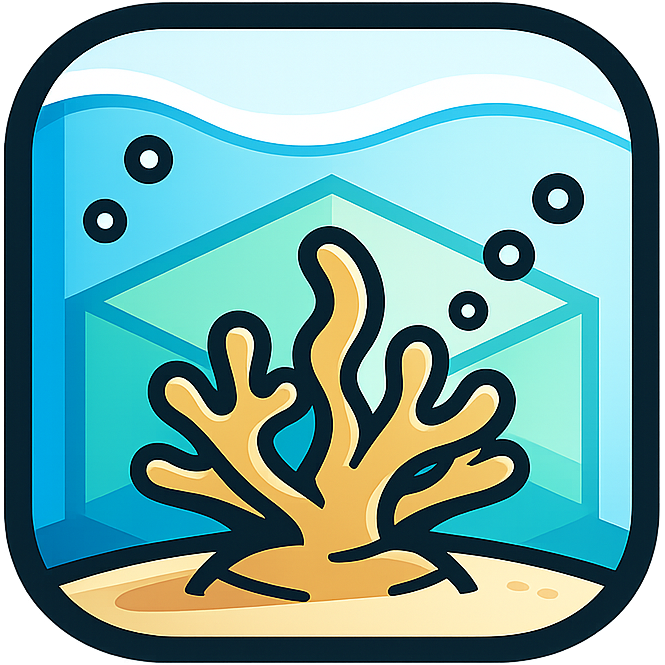}}\hspace{0.35em}}{}3D-USE: From Image-Level to Scene-Level Underwater Enhancement}

\author{
Jieyu Yuan$^{1}$, Yuanlin Zhang$^{1}$, Jihong Li$^{1}$,
Chunle Guo$^{1,2}$, Huimin Lu$^{3,4}$, Chongyi Li$^{1,2,*}$\\
{\small $^{1}$College of Computer Science, Nankai University;\quad $^{2}$NKIARI, Shenzhen Futian}\\
{\small $^{3}$School of Automation, Southeast University;\quad $^{4}$Advanced Ocean Institute of Southeast University, Nantong}\\
{\small \texttt{jieyuyuan.cn@gmail.com}, \texttt{lichongyi@nankai.edu.cn}}
}

\makeatletter
\newcommand{\arxivmaketitle}{%
  \begingroup
  \patchcmd{\@maketitle}{\vskip .375in}{\vskip 0pt}{}{}%
  \patchcmd{\@maketitle}{\vspace*{12pt}}{}{}{}%
  \patchcmd{\@maketitle}{\vspace*{24pt}}{\vspace*{12pt}}{}{}%
  \patchcmd{\@maketitle}{\vskip .5em}{}{}{}%
  \maketitle
  \endgroup
}
\makeatother

\renewcommand{\maketitlesupplementary}{%
  \twocolumn[{%
    \centering
    {\Large\bfseries\thetitle\\[0.4em]
      {\normalsize\itshape\rule[0.45ex]{1.7cm}{0.4pt}\enspace Supplementary Material\enspace\rule[0.45ex]{1.7cm}{0.4pt}}\par}
    \vspace{1em}%
  }]%
}

\begin{document}
\input{sec/teaser}
\begingroup
\patchcmd{\abstract}{\vspace*{12pt}}{\vspace*{6pt}}{}{}
\patchcmd{\endabstract}{\vspace*{12pt}}{}{}{}
\input{sec/0_abstract}
\endgroup
\input{sec/1_intro}
\input{sec/2_related_work}
\input{sec/3_method}
\input{sec/4_experiments}
\input{sec/5_conclusion}
\newpage

\section{LLM Usage Statement}
We used OpenAI Codex to assist with manuscript revision, the creation of illustrative icons and schematic elements, and the implementation of a web-based interface for supplementary video comparisons. All AI-assisted content was reviewed by the authors, who take full responsibility for the methodology, experimental results, and conclusions.

{
    \small
    \bibliographystyle{ieeenat_fullname}
    \bibliography{main}
}

\input{sec/X_suppl}

\end{document}

%% file: preamble.tex
\usepackage{cuted}%

\usepackage{currfile}%

\usepackage{caption}%

\input{project_packages}

%% file: project_packages.tex
\usepackage{placeins}
\makeatletter
\renewcommand{\LenToUnit}[1]{\strip@pt\dimexpr#1\relax}
\makeatother

\AtBeginDocument{%
  \crefname{figure}{Figure}{Figures}%
  \Crefname{figure}{Figure}{Figures}%
}

%% file: sec/teaser.tex
\twocolumn[{%
	\renewcommand\twocolumn[1][]{#1}%
	\arxivmaketitle
	{\centering
		\captionsetup{type=figure}
		\includegraphics[width=1\textwidth]{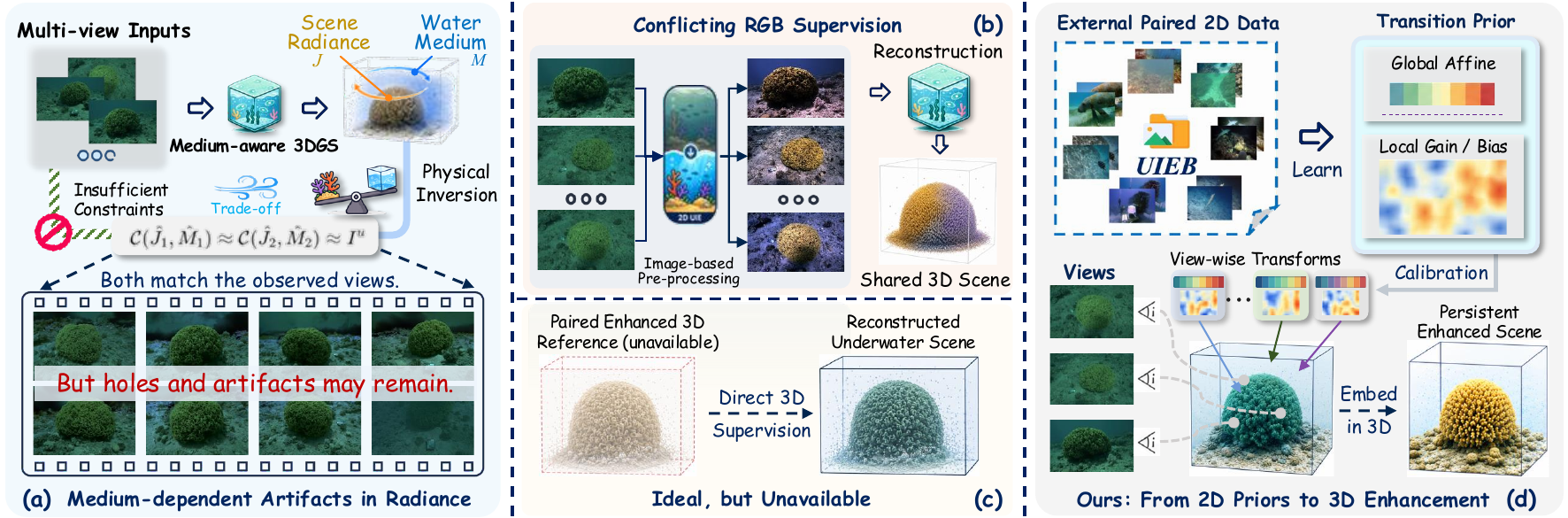}
		\vspace{-0.22in}
		\captionof{figure}{
       \textbf{Motivation of 3D-USE.} (a) Different scene radiance and medium estimates can match the same observed views, making physical inversion ambiguous and allowing color and spatial artifacts to remain in recovered radiance (bottom renderings \cite{li2025watersplatting}). (b) Independent 2D enhancement may give the same object different colors across views, creating conflicting RGB targets when reconstructing a shared 3D scene. (c) Paired enhanced 3D references would provide direct supervision but are difficult to obtain. (d) We \textbf{transfer appearance transitions learned from external paired 2D data} into shared scene-level guidance, enabling consistent enhanced novel-view rendering.
	}
		\label{fig:teaser}
	\par\medskip}%
}]
{\renewcommand{\thefootnote}{\fnsymbol{footnote}}\footnotetext[1]{Corresponding author.}}

%% file: sec/0_abstract.tex
\begin{abstract}
    Underwater scene reconstruction from degraded views can preserve visual degradation. Since paired enhanced 3D references are hard to obtain in real ocean scenes, underwater 3D methods largely rely on physics-based restoration. However, separating object radiance from medium effects is ill-posed and may leave residual artifacts. In contrast, we formulate underwater scene enhancement as learning the desired appearance of the scene and propose a two-stage framework called 3D-USE. Stage~1 builds a medium-aware Gaussian scene with MediumRBF, sharing medium estimates across nearby views and establishing stable geometry and visibility. Using a prior learned from external 2D enhancement pairs, Stage~2 applies Appearance Transition Consensus (ATC) to calibrate and aggregate view-wise appearance changes, defining the desired scene-level appearance transformation. Finally, an Underwater Bilateral Appearance Field (U-BAF) jointly adjusts object radiance and medium appearance to produce coherent enhanced appearance across viewpoints. Experiments show that 3D-USE improves novel-view enhancement and cross-view consistency while maintaining competitive reconstruction \mbox{quality. The project page and code are available \href{https://bilityniu.github.io/3D-USE/}{here}.}
\par
\end{abstract}

%% file: sec/1_intro.tex
\section{Introduction}
Constructing underwater 3D scenes with good visibility is important for marine exploration and analysis. However, wavelength-dependent attenuation and path-dependent backscatter cause color casts and contrast loss~\cite{akkaynak2017space,akkaynak2018revised}. Although 3D Gaussian Splatting (3DGS) enables high-quality reconstruction~\cite{3DGS_tog_2023,2dgs,Mip-Splatting_2024_CVPR}, faithfully reproducing underwater observations can also preserve their degradation in the reconstructed scene. Therefore, we aim to construct an enhanced 3D scene that directly renders novel views with improved visibility and consistent appearance.

Existing underwater reconstruction methods estimate and compensate for attenuation and scattering to recover scene appearance with reduced medium effects~\cite{li2025watersplatting,UW-GS_WACV_2025,RestorGS_cvpr_2025}. Separating scene radiance from transmission and backscatter is ambiguous, as different combinations can recover the same observed colors~\cite{WaterNeRF_2023}. Across viewpoints, the same scene content is observed through different paths, with varying transmission and backscatter~\cite{wu2025plenodium}. Multi-view observations help constrain this separation~\cite{boittiaux2024sucre}, but changes in scene radiance still compensate for errors in medium estimation. The restored appearance may therefore retain color distortions and spatial artifacts, as shown in Figure~\ref{fig:teaser}(a). Removing estimated water effects alone does not necessarily provide the color and visibility desired for scene enhancement.

Data-driven underwater image enhancement (UIE) learns appearance mappings between degraded images and enhanced references to directly improve color and contrast~\cite{UIEB_TIP_2019,PUGAN_TIP_2023,LSUI_TIP_2023}. This motivates us to formulate scene enhancement as a distinct learning objective, rather than directly treating the radiance recovered through physical inversion as the enhanced result. However, extending 2D enhancement capabilities to 3D scenes is not straightforward: the same scene content may receive incompatible appearance corrections across viewpoints, imposing conflicting RGB supervision on a shared 3D representation~\cite{Liu_2025_CVPR,cui_luminance_gs,SplatSuRe_cvpr_2026} (Figure~\ref{fig:teaser}(b)). Such view-dependent variation also motivates consistency-aware enhancement of underwater image sequences~\cite{xie2024uveb, boittiaux2024sucre}. Meanwhile, paired enhanced 3D references that could provide direct supervision are difficult to obtain in practice (Figure~\ref{fig:teaser}(c)).%
The central challenge is to determine a coherent enhanced appearance for the entire 3D scene from 2D enhancement knowledge, rather than relying on potentially conflicting view-wise RGB targets.

Our goal is to improve underwater scene visibility by learning the desired enhanced appearance, rather than directly adopting radiance estimates from ill-posed physical inversion. To this end, we propose \textbf{3D-USE}, a two-stage framework shown in Figure~\ref{fig:teaser}(d). Stage~1 reconstructs a Gaussian scene, providing geometry and visibility for combining corrections from different views on the same 3D content. To reduce water-induced bias in this reconstruction, MediumRBF models the medium separately from object radiance using adaptive radial-basis anchors shared across nearby viewpoints. To specify the desired appearance without enhanced references for the target scene, we learn an appearance-transition prior offline from external pairs of underwater images and enhanced references. In Stage~2, Appearance Transition Consensus (ATC) calibrates view-wise proposals and combines them into shared global and local targets to coordinate corrections across views. Since both object radiance and medium effects contribute to the rendered image, an Underwater Bilateral Appearance Field (U-BAF) learns additional transformations of both components while leaving the Stage~1 parameters unchanged. Supervision on the resulting appearance changes constrains their combined effect in the rendered image. Together, these designs transfer image-level enhancement knowledge into a shared 3D scene that directly renders enhanced novel views, enabling \textbf{Underwater Scene-level Enhancement (USE)} without paired enhanced 3D supervision. Experiments on real underwater scenes show improved visibility, fewer local color artifacts, and better average cross-view consistency than the compared reconstruction-based methods.

Our main contributions are summarized as follows:
\begin{itemize}
    \item We propose \textbf{3D-USE} to improve underwater scene visibility by learning a coherent enhanced appearance within reconstructed 3D geometry. This design reduces color casts and local artifacts from physical restoration while improving cross-view appearance consistency.
    \item We introduce appearance-transition supervision that uses knowledge learned from paired 2D data to construct shared global and local targets, establishing scene-level enhancement guidance in the absence of paired enhanced 3D references.
    \item We design \textbf{MediumRBF}, which uses adaptive radial-basis anchors to share medium information across nearby viewpoints. It models water effects separately from Gaussian radiance, reducing medium-related artifacts and providing a more reliable basis for scene enhancement.
\end{itemize}

%% file: sec/2_related_work.tex
\section{Related Work}

\noindent\textbf{Underwater Medium Modeling.} Underwater neural rendering jointly models scene radiance and light propagation through water. SeaThru-NeRF incorporates medium effects into NeRF~\cite{levy2023seathrunerf}, while 3DGS-based methods model attenuation and backscatter during Gaussian splatting for efficient underwater reconstruction~\cite{li2025watersplatting,UW-GS_WACV_2025,RestorGS_cvpr_2025,yuan2025_3duir}. Plenodium further represents spatial and directional variations in the medium~\cite{wu2025plenodium}, while other approaches reduce medium-induced reconstruction artifacts~\cite{kweon2026oceansplat,marinestd_2025_acmmm}.
However, errors in medium estimation can leave color casts and haze in the recovered radiance even when rendered views closely match the captured images. Therefore, we retain physical modeling for reconstruction and learn the desired enhanced appearance separately, rather than directly treating the recovered radiance as the enhanced result.

\begin{figure*}[ht]
	\centering
	\includegraphics[width=1\linewidth]{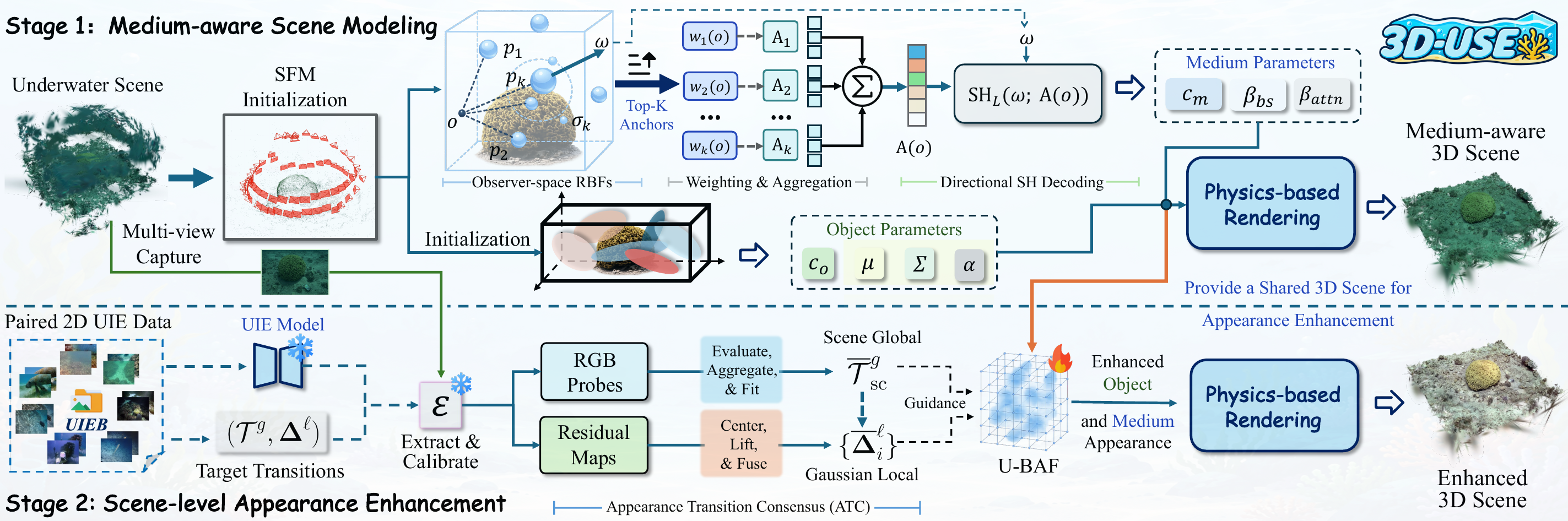}
	\caption{
		\textbf{Overview of 3D-USE.}
        Stage~1 reconstructs a medium-aware Gaussian scene, with MediumRBF modeling medium variations. An appearance-transition prior is learned separately from paired 2D UIE data. With Stage~1 fixed, Stage~2 uses ATC to convert calibrated view-wise transitions into scene-global and Gaussian-local targets that guide U-BAF to jointly enhance object and medium appearance.
	}
	\label{fig:pipeline}
\end{figure*}

\noindent\textbf{Underwater Image Enhancement.} Underwater image enhancement (UIE) improves color, contrast, and visibility through physical modeling and data-driven learning~\cite{seathru_cvpr_2019,UIEB_TIP_2019,Ucolor_TIP_2021,PUGAN_TIP_2023,uranker_aaai_2023,yuan_JOE_2025}. Recent methods exploit generative models and diffusion priors for perceptual enhancement and restoration~\cite{osmosis_eccv_2024,slurpp_tpami_2025,W2WDiff_tgrs_2026}, while video enhancement incorporates temporal information to improve consistency across frames~\cite{xie2024uveb}. However, independently enhanced training views can provide inconsistent supervision for a shared 3D representation~\cite{WaterHENeRF_2024_IF}. We transfer 2D enhancement knowledge through appearance transitions rather than direct supervision with enhanced RGB targets.

\noindent\textbf{Scene-Level Appearance Control.} Neural scene representations provide a shared 3D basis for consolidating appearance cues across viewpoints~\cite{Plenoxels_2022_CVPR,matspray_cvpr_2026,intergsedit_iccv_2025}. NeRF-W and subsequent 3DGS methods model photometric variations across input images using per-image embeddings or learned appearance features~\cite{martinbrualla2021nerfw,dahmani2024swag,zhang2024GS-W}. These methods primarily account for appearance variations in the observations rather than define an enhanced scene appearance. BilaRF instead transfers a prescribed 2D appearance edit to a shared 3D representation for consistent rendering across viewpoints~\cite{BilRF_tog_2024}. These methods either model appearance variations in observations or rely on a specified edit. In contrast, underwater scene enhancement requires improving color and visibility while preserving scene content and cross-view consistency, without a prescribed target appearance. We learn appearance transitions from paired UIE images to guide coherent 3D scene enhancement.

%% file: sec/3_method.tex
\section{Methodology}

As shown in Figure~\ref{fig:pipeline}, Stage~1 jointly estimates Gaussian geometry, object radiance, and medium parameters to reconstruct the captured underwater views (Sec.~\ref{sec:medium_rbf}). Separately, we learn an appearance-transition prior offline from external paired 2D data and reuse it across scenes. With the reconstruction fixed, Stage~2 applies this frozen prior to estimate appearance changes for each view and combines them into a global affine target and local targets associated with the Gaussians (Sec.~\ref{sec:atc}). Guided by these targets, we learn coordinated changes to object radiance and medium appearance through a shared appearance field to produce enhanced renderings (Sec.~\ref{sec:ubaf}). The optimization objectives for both stages are detailed in Sec.~\ref{sec:optimization}.

\subsection{Medium Radial Basis Anchor Representation}
\label{sec:medium_rbf}

Underwater observations combine object radiance with medium effects that vary with observer position and ray direction. To reduce their coupling, we design MediumRBF to capture local medium variations while sharing information across nearby viewpoints. The representation uses learnable medium anchors with radial-basis spatial weighting~\cite{NeuRBF_2023_ICCV} and spherical-harmonic (SH) coefficients for directional modeling. Learnable centers and widths control each anchor's spatial influence, while Top-$K$ selection restricts aggregation to the most relevant anchors. The aggregated SH coefficients are then decoded along each ray direction to obtain the medium parameters.

For a camera ray $\mathbf r=(\mathbf o,\boldsymbol\omega)$, we represent the medium using a set of learnable anchors. Each anchor stores a center $\mathbf p_k$, a positive support width $\sigma_k$, and SH coefficients $\mathbf A_k$. The unnormalized affinity score $\rho_k(\mathbf o)$ of anchor $k$ at observer position $\mathbf o$ is
\begin{equation}
    \rho_k(\mathbf o)=-\frac{\lVert\mathbf o-\mathbf p_k\rVert_2^2}{2\sigma_k^2}.
    \label{eq:medium_rbf_score}
\end{equation}

Let $\mathcal K(\mathbf o)$ contain the $K$ anchors with the highest scores. The normalized weights and aggregated coefficients are
\begin{equation}
    \begin{aligned}
        w_k(\mathbf o)&=\frac{\exp(\rho_k(\mathbf o))}{\sum_{j\in\mathcal K(\mathbf o)}\exp(\rho_j(\mathbf o))},\\
        \mathbf A(\mathbf o)&=\sum_{k\in\mathcal K(\mathbf o)}w_k(\mathbf o)\mathbf A_k.
    \end{aligned}
    \label{eq:medium_rbf_weight}
\end{equation}
The aggregated coefficients define the medium parameters along the unit ray direction $\boldsymbol\omega$ as
\begin{equation}
    [\mathbf c_m,\boldsymbol\beta_{\mathrm{bs}},\boldsymbol\beta_{\mathrm{attn}}]=\Phi\!\left(\operatorname{SH}_{L}(\boldsymbol\omega;\mathbf A(\mathbf o))\right).
    \label{eq:medium_rbf_decode}
\end{equation}
Here, $\operatorname{SH}_{L}$ evaluates degree-$L$ SH, and $\Phi$ constrains the outputs to be non-negative. The outputs are the asymptotic medium color $\mathbf c_m$ and the RGB backscatter and attenuation coefficients $\boldsymbol\beta_{\mathrm{bs}}$ and $\boldsymbol\beta_{\mathrm{attn}}$. These effective parameters are treated as constant along each ray.

We incorporate the decoded medium parameters into underwater compositing~\cite{li2025watersplatting,wu2025plenodium} and jointly optimize the anchors, Gaussian geometry, and object radiance to reconstruct the captured views. Establishing the geometry and visibility first allows corrections from different views to be combined on the same 3D content. We retain this reconstruction as a fixed reference for learning the enhanced appearance without changing the scene structure.

\subsection{Enhancement Target Construction}
\label{sec:atc}

To define how the reconstructed scene should be enhanced, we use external paired underwater and enhanced images~\cite{UIEB_TIP_2019} as references for the desired enhanced appearance. Appearance Transition Consensus (ATC) transfers this prior to the target scene through a learned calibrator and combines view-wise proposals into a shared global transformation and local targets for individual Gaussians.

We train the calibrator offline to adjust the changes proposed by a 2D enhancer toward those observed in the paired references. A differentiable extractor $\mathcal E$ describes the change between two images using a global affine operator and a grid of local RGB gains and biases. For a pair $(\mathbf I^u,\mathbf I^e)$, $\mathcal E(\mathbf I^u,\mathbf I^e)$ defines the reference transition, while $\mathcal E(\mathbf I^u,\mathcal N(\mathbf I^u))$ gives the proposal from a frozen UIE model $\mathcal N$~\cite{uranker_aaai_2023}. Conditioned on source-image descriptors, the calibrator learns to map the proposal toward the reference transition. The extractor, UIE model, and calibrator remain frozen during scene optimization.

For a target scene, the calibrated transitions still differ across views. Therefore, we combine their global components into a shared color transformation. Specifically, the calibrator predicts a candidate $\widehat{\mathcal T}_v^g$ for each captured image $\mathbf I_v^u$. We evaluate these candidates on a common RGB probe chart, take the geometric median of their responses, and fit a single affine operator $\overline{\mathcal T}_{\mathrm{sc}}^g$ by least squares.

Since a global transformation cannot capture all spatial variations, we further construct local correction targets. Conditioned on $\overline{\mathcal T}_{\mathrm{sc}}^g$, the calibrator's local head predicts a residual map for each view. Subtracting its spatial mean gives $\widehat{\boldsymbol\Delta}_v^\ell$, separating local corrections from the global transformation. To associate these image-level residuals with the shared 3D scene, we map them to Gaussians using the fixed splatting weights. For Gaussian $i$ visible in view $v$, let $w_{iv}(p)$ denote its splatting weight at pixel $p$. The corresponding residual is
\begin{equation}
    \widehat{\boldsymbol\Delta}_{iv}^{\ell}
    =\frac{\sum_p [w_{iv}(p)\,\widehat{\boldsymbol\Delta}_v^{\ell}(p)]}{\sum_p w_{iv}(p) +\epsilon}.
    \label{eq:local_lifting}
\end{equation}
Here, $\widehat{\boldsymbol\Delta}_v^\ell(p)$ is the residual sampled at pixel $p$, and $\epsilon$ ensures numerical stability.%
Visibility-weighted robust fusion combines the residuals from different views into one target $\overline{\boldsymbol\Delta}_i^\ell$ for each Gaussian. Together with the scene-global operator, these targets provide fixed supervision for Stage~2.

\subsection{Scene Appearance Transformation}
\label{sec:ubaf}
Guided by the ATC targets, we learn an appearance mapping that transforms Gaussian radiance and medium appearance while keeping the Stage~1 reconstruction fixed. U-BAF predicts local color corrections through a shared bilateral field~\cite{gharbi2017deepbilateral,BilRF_tog_2024}. The field is indexed by 3D position and a scalar guide $\gamma(\bar{\mathbf c}_i)$ computed from the fixed, view-independent Stage~1 base color $\bar{\mathbf c}_i$ through a learned guide function. The field uses a low-rank representation to predict local RGB log-gains and biases.

Let $F_\theta$ denote the local field and $\mathcal T_\theta^g(\mathbf c)=\mathbf A_\theta\mathbf c+\mathbf b_\theta$ the trainable scene-global affine. For Gaussian $i$ at position $\mathbf x_i$, the enhanced radiance is
\begin{equation}
    \begin{aligned}
        (\mathbf a_i^\ell,\mathbf b_i^\ell)&=F_\theta(\mathbf x_i,\gamma(\bar{\mathbf c}_i)),\\
        \mathbf c_i^+(\boldsymbol\omega)&=\exp(\mathbf a_i^\ell)\odot\mathcal T_\theta^g\!\left(\mathbf c_i(\boldsymbol\omega)\right)+\mathbf b_i^\ell.
    \end{aligned}
    \label{eq:enhanced_gaussian}
\end{equation}
Here, $\odot$ denotes element-wise multiplication. $F_\theta$ returns local RGB log-gains $\mathbf a_i^\ell$ and biases $\mathbf b_i^\ell$, centered over the Gaussians to reduce overlap with the global affine.

The same field and global affine also transform medium color, with the field evaluated at the camera position using a guide derived from the MediumRBF color. Attenuation and backscatter are independently scaled by bounded positive factors shared across the scene, preserving their Stage~1 RGB channel proportions. The underwater compositor combines the transformed Gaussian radiance and medium parameters to produce the enhanced rendering $\mathbf I_v^+$.

\begin{figure*}[th]
	\centering
	\includegraphics[width=0.96\linewidth]{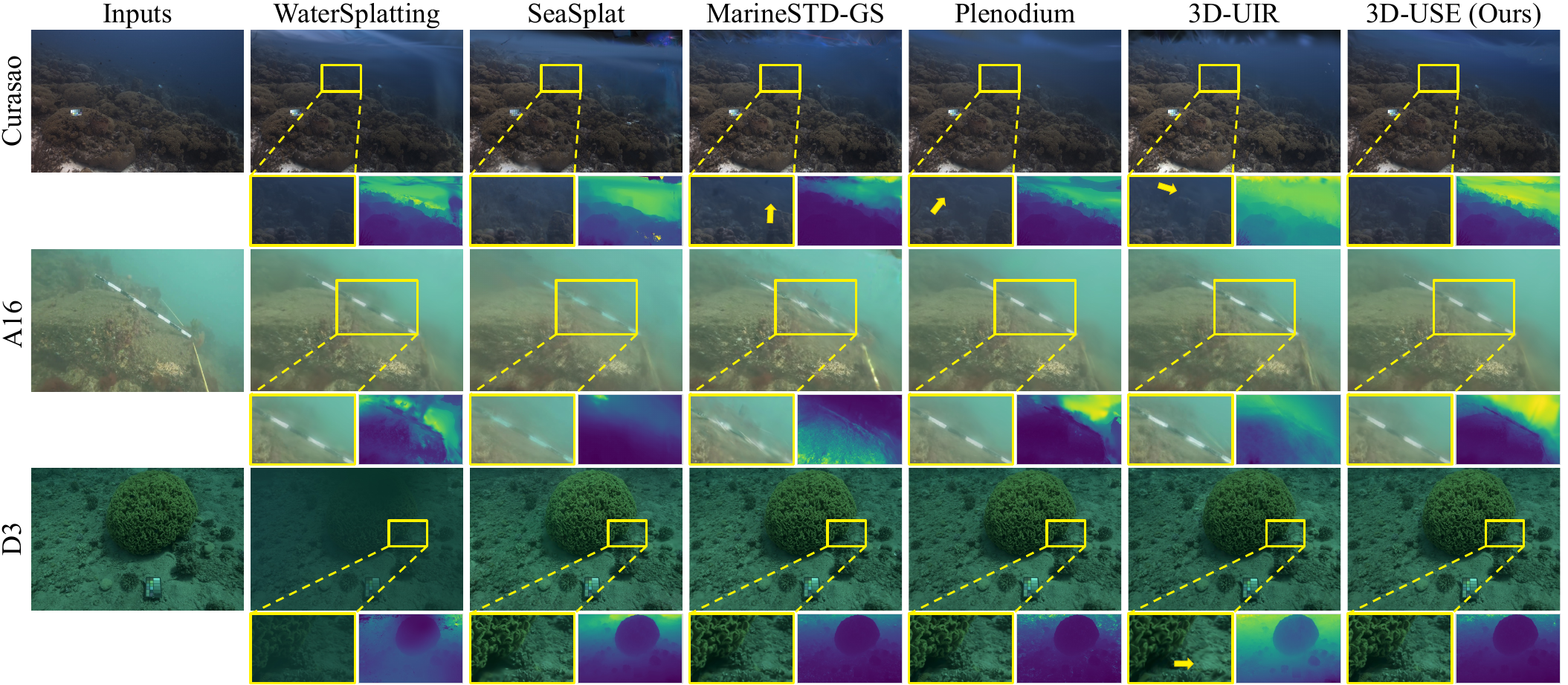}
    \vspace{-5pt}
	\caption{Qualitative comparison of underwater scene reconstruction. Zoomed regions highlight background consistency and medium-related artifacts. 3D-USE produces cleaner scene appearance with fewer medium-induced inconsistencies.}
	\label{fig:stage1_restlt}
\end{figure*}

\begin{table*}[h]
	\caption{Quantitative comparison of reconstruction quality (PSNR$\uparrow$, SSIM$\uparrow$, and LPIPS$\downarrow$) and rendering speed (FPS$\uparrow$). SeaThru-4 and DRUVA-20 report averages across scenes. FPS is averaged across scenes. Best and second-best results are bold and underlined.}
    \vspace{-5pt}
	\label{tab:reconstruction}
	\centering
	\small
	\setlength{\tabcolsep}{0.90mm}
	\begin{tabular}{@{}l*{12}{c}c@{}}
		\toprule
		& \multicolumn{3}{c}{SeaThru-4}
		& \multicolumn{3}{c}{DRUVA-20}
		& \multicolumn{3}{c}{D3}
		& \multicolumn{3}{c}{D5}
		& Avg. \\
		\cmidrule(lr){2-4}\cmidrule(lr){5-7}\cmidrule(lr){8-10}\cmidrule(lr){11-13}\cmidrule(lr){14-14}
		Method
		& PSNR & SSIM & LPIPS
		& PSNR & SSIM & LPIPS
		& PSNR & SSIM & LPIPS
		& PSNR & SSIM & LPIPS
		& FPS \\
		\midrule
		\shortstack[l]{WaterSplatting (3DV, \citeyear{li2025watersplatting})}
		& 29.340 & 0.919 & 0.129
		& 27.007 & 0.864 & 0.357
		& 27.380 & 0.776 & 0.253
		& 29.113 & 0.817 & \underline{0.313}
		& 99.78 \\
		\shortstack[l]{SeaSplat (ICRA, \citeyear{yang2025seasplat})}
		& 27.146 & 0.878 & 0.184
		& 22.866 & 0.828 & 0.388
		& 27.885 & 0.785 & 0.282
		& 27.113 & 0.806 & 0.382
		& \underline{100.77} \\
		\shortstack[l]{MarineSTD-GS (ACM MM, \citeyear{marinestd_2025_acmmm})}
		& 30.034 & 0.901 & 0.160
		& 26.472 & 0.856 & 0.364
		& \underline{29.510} & 0.781 & 0.316
		& \underline{30.237} & 0.818 & 0.329
		& 41.13 \\
		\shortstack[l]{Plenodium (NeurIPS, \citeyear{wu2025plenodium})}
		& \underline{30.331} & \underline{0.922} & \underline{0.125}
		& \underline{27.413} & \underline{0.866} & \textbf{0.335}
		& 28.686 & 0.803 & \underline{0.218}
		& 29.088 & 0.817 & 0.314
		& 91.60 \\
		\shortstack[l]{3D-UIR (IEEE TIP, \citeyear{yuan2025_3duir})}
		& 28.260 & 0.889 & 0.179
		& 25.432 & 0.820 & 0.354
		& 28.098 & \underline{0.830} & 0.283
		& 28.973 & \underline{0.823} & 0.324
		& 42.74 \\
		\midrule
		\textbf{3D-USE (Ours)}
		& \textbf{30.369} & \textbf{0.924} & \textbf{0.124}
		& \textbf{27.495} & \textbf{0.867} & \underline{0.352}
		& \textbf{29.759} & \textbf{0.858} & \textbf{0.183}
		& \textbf{31.400} & \textbf{0.908} & \textbf{0.235}
		& \textbf{105.36} \\
		\bottomrule
	\end{tabular}

\end{table*}

\subsection{Optimization Objectives}
\label{sec:optimization}

\noindent\textbf{Stage~1.} Gaussian geometry, object radiance, and MediumRBF parameters are jointly optimized to reconstruct the captured underwater views:
\begin{equation}
    \mathcal L_{\mathrm{S1}}=\mathcal L_{\mathrm{rec}}+\lambda_d\mathcal L_{\mathrm{depth}}.
    \label{eq:stage1_objective}
\end{equation}
The $\mathcal L_{\mathrm{rec}}$ term combines $\ell_1$ and MS-SSIM losses computed on images weighted by the inverse intensity of the detached rendering~\cite{Mildenhall_2022_CVPR, LE3D_2024, wu2025plenodium}. The $\mathcal L_{\mathrm{depth}}$ term is one minus the Pearson correlation between rendered disparity and the Depth Anything V2 prior~\cite{Depth_Anything_V2}.

\noindent\textbf{Stage~2.} We learn transformations of object radiance and medium appearance by matching the appearance changes between Stage~1 and enhanced renderings to the ATC targets. The complete objective is
\begin{equation}
    \mathcal L_{\mathrm{S2}}=\lambda_g\mathcal L_{\mathrm{global}}+\lambda_\ell\mathcal L_{\mathrm{local}}+\lambda_{\mathrm{tv}}\mathcal L_{\mathrm{TV}}.
    \label{eq:stage2_objective}
\end{equation}
Here, $\mathcal L_{\mathrm{TV}}$ regularizes the U-BAF factor grids. To compute the transition losses for view $v$, we apply the extractor $\mathcal E$ to the Stage~1 rendering $\mathbf I_v^{\mathrm{rec}}$ and enhanced rendering $\mathbf I_v^+$:
\begin{equation}
    (\mathbf t_v^g,\boldsymbol\Delta_v^\ell)=\mathcal E\!\left(\operatorname{sg}(\mathbf I_v^{\mathrm{rec}}),\mathbf I_v^+\right),
    \label{eq:stage2_realized_transition}
\end{equation}
where $\operatorname{sg}(\cdot)$ stops gradients. Global affine parameters $\mathbf t_v^g$ and local gain-and-bias map $\boldsymbol\Delta_v^\ell$ describe the combined effect of object and medium adjustments in the rendered view.

Subtracting the spatial mean of $\boldsymbol\Delta_v^\ell$ gives the centered residual $\widetilde{\boldsymbol\Delta}_v^\ell$. Splatting the Gaussian targets $\{\overline{\boldsymbol\Delta}_i^\ell\}$ constructed by ATC yields the view-space target $\overline{\boldsymbol\Delta}_v^\ell$, aligned with the extracted residual map. Let $\overline{\mathbf t}_{\mathrm{sc}}^g$ denote the parameter vector of the ATC global operator $\overline{\mathcal T}_{\mathrm{sc}}^g$. The matching losses are
\begin{equation}
    \begin{aligned}
        \mathcal L_{\mathrm{global}}&=\left\|\mathbf t_v^g-\overline{\mathbf t}_{\mathrm{sc}}^g\right\|_{\mathrm{S1}},\\
        \mathcal L_{\mathrm{local}}&=\left\|\widetilde{\boldsymbol\Delta}_v^\ell-\overline{\boldsymbol\Delta}_v^\ell\right\|_{\mathrm{S1}}.
    \end{aligned}
    \label{eq:stage2_transition_losses}
\end{equation}
Here, $\|\cdot\|_{\mathrm{S1}}$ denotes the reduced Smooth-L1 penalty. Both terms use component-wise normalization with precomputed scales, and the local term also uses normalized visibility weights over the residual grid. Stage~2 updates only the appearance transformation parameters, leaving the Stage~1 model parameters and ATC targets unchanged. Further details are provided in the \textcolor{blue}{supplementary material}.

\begin{figure*}[t]
	\centering
	\includegraphics[width=1\linewidth]{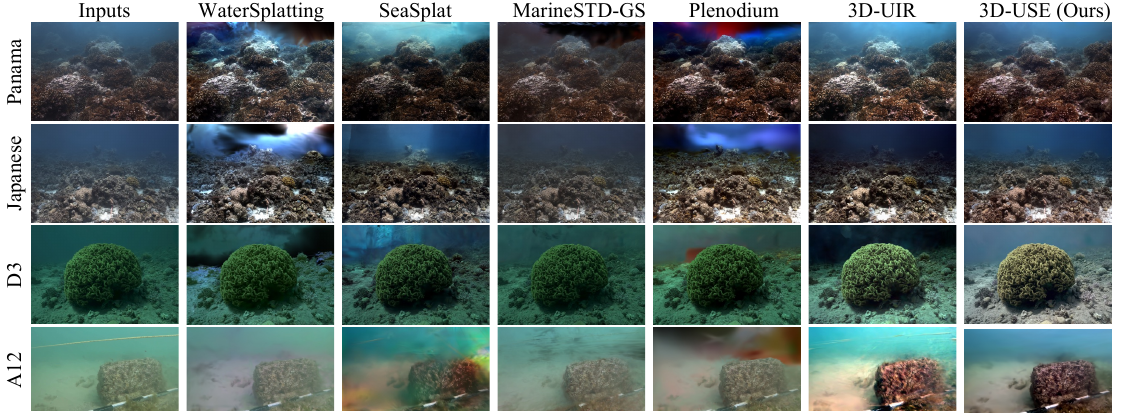}
    \vspace{-5pt}
	\caption{Qualitative comparison of enhanced novel-view renderings across different underwater scenes. Compared with existing methods, 3D-USE achieves more balanced color correction and visibility enhancement with fewer spatial artifacts.}
	\label{fig:stage2_restlt}
\end{figure*}

\begin{table*}[t]
	\caption{Enhancement quality (UCIQE$\uparrow$, URanker$\uparrow$, MUSIQ$\uparrow$) and cross-view consistency (wLPIPS$\downarrow$). SeaThru-4 and DRUVA-20 use scene-macro averages; wLPIPS averages all 26 scenes equally. Best/second-best results are bold/underlined.}
    \vspace{-5pt}
	\label{tab:enhancement}
	\centering
	\small
	\setlength{\tabcolsep}{0.69mm}
	\begin{tabular}{@{}l*{13}{c}@{}}
		\toprule
		& \multicolumn{3}{c}{SeaThru-4}
		& \multicolumn{3}{c}{DRUVA-20}
		& \multicolumn{3}{c}{D3}
		& \multicolumn{3}{c}{D5}
		& Avg. \\
		\cmidrule(lr){2-4}\cmidrule(lr){5-7}\cmidrule(lr){8-10}\cmidrule(lr){11-13}\cmidrule(lr){14-14}
		Method
		& UCIQE & URanker & MUSIQ
		& UCIQE & URanker & MUSIQ
		& UCIQE & URanker & MUSIQ
		& UCIQE & URanker & MUSIQ
		& wLPIPS \\
		\midrule
		WaterSplatting
		& 0.508 & 1.361 & \underline{58.529}
		& 0.385 & $-$0.856 & 33.193
		& 0.470 & 0.541 & 58.833
		& \underline{0.499} & \underline{0.690} & \underline{55.568}
		& 0.1639 \\
		SeaSplat
		& 0.588 & 1.555 & 54.470
		& 0.518 & $-$0.186 & 28.252
		& 0.504 & 1.007 & 58.251
		& 0.442 & $-$0.895 & 40.648
		& 0.2211 \\
		Plenodium
		& 0.503 & 1.058 & 58.370
		& 0.395 & $-$0.792 & 34.786
		& 0.483 & 0.705 & \underline{60.158}
		& 0.443 & 0.033 & 52.456
		& \underline{0.1481} \\
		MarineSTD-GS
		& 0.522 & 0.918 & 55.890
		& 0.447 & $-$0.567 & \textbf{39.070}
		& 0.443 & 0.416 & 56.246
		& 0.413 & $-$0.117 & 54.424
		& 0.2100 \\
		3D-UIR
		& \underline{0.609} & \underline{1.658} & 54.540
		& \underline{0.537} & \underline{0.174} & 32.752
		& \underline{0.519} & \underline{1.356} & 59.351
		& 0.468 & 0.137 & 51.991
		& 0.2228 \\
		\midrule
		\textbf{3D-USE (Ours)}
		& \textbf{0.609} & \textbf{1.679} & \textbf{59.881}
		& \textbf{0.575} & \textbf{0.456} & \underline{36.598}
		& \textbf{0.543} & \textbf{2.034} & \textbf{62.993}
		& \textbf{0.513} & \textbf{1.027} & \textbf{58.324}
		& \textbf{0.1381} \\
		\bottomrule
	\end{tabular}

\end{table*}

%% file: sec/4_experiments.tex
\section{Experiments}
\subsection{Evaluation Protocol}
\noindent\textbf{Datasets.}
We evaluate our method on four real-world scenes from SeaThru-NeRF~\cite{levy2023seathrunerf}, namely IUI3-RedSea, Curasao, Japanese Gardens, and Panama, together with the D3 and D5 sequences from Sea-Thru~\cite{seathru_cvpr_2019}. All six scenes are evaluated at $1400\times933$ resolution. We further use all 20 scenes from DRUVA~\cite{DRUVA_ICCV_2023} at $1920\times1080$ resolution. These datasets cover diverse water conditions, scene contents, and camera trajectories. The paired UIE data from UIEB~\cite{UIEB_TIP_2019} used to learn the transition prior are disjoint from all target 3D scenes and their test views.

\noindent\textbf{Evaluation Metrics.}
We evaluate reconstruction quality using PSNR, SSIM, and LPIPS. Since paired enhanced 3D ground truth is unavailable, we follow previous UIE works~\cite{Huang_2023_CVPR,osmosis_eccv_2024,yuan2025_3duir} and assess enhanced novel views using the no-reference underwater metrics  UCIQE~\cite{uciqe_TIP_2015} and URanker~\cite{uranker_aaai_2023}, complemented by MUSIQ~\cite{MUSIQ_iccv_2021} for general perceptual quality. We further measure cross-view appearance consistency using warped LPIPS (wLPIPS)~\cite{stylizednerf_cvpr_2022}. Rendering speed is reported in FPS at a resolution of $1400\times933$.

\noindent\textbf{Baselines.}
We compare our method with WaterSplatting~\cite{li2025watersplatting}, SeaSplat~\cite{yang2025seasplat}, MarineSTD-GS~\cite{marinestd_2025_acmmm}, Plenodium~\cite{wu2025plenodium}, and 3D-UIR~\cite{yuan2025_3duir}. For scene enhancement, we evaluate the enhanced outputs defined by each method.%

\noindent\textbf{Implementation Details.}
Our method is implemented in Nerfstudio~\cite{nerfstudio} and uses gsplat~\cite{ye2025gsplat} for differentiable rasterization. The 3D Gaussians are initialized from sparse point clouds reconstructed by COLMAP~\cite{SFM_2016_CVPR}. All experiments are conducted on a single NVIDIA RTX 3090 GPU.%

\subsection{Experimental Results}

\noindent\textbf{Reconstruction Quality.}
Table~\ref{tab:reconstruction} reports reconstruction performance, and the complete-pipeline rendering speed is measured at $1400\times933$ resolution. 3D-USE achieves the best PSNR and SSIM on all four benchmarks while maintaining competitive LPIPS and the highest rendering speed. The gains are particularly notable on D3 and D5, where 3D-USE improves PSNR by 0.249 and 1.163 dB over the second-best methods, respectively.

\begin{table}[t]
    \caption{Training time and rendering speed (our timings: SeaThru-4, one RTX~3090). Total includes both stages. Best and second-best complete methods are bold and underlined.}
    \vspace{-5pt}
    \label{tab:efficiency}
    \centering
    \small
    \begin{tabular}{
        @{}l
        @{\hspace{18pt}}c
        @{\hspace{18pt}}c@{}
    }
        \toprule
        Method & Time (min)$\downarrow$ & FPS$\uparrow$ \\
        \midrule
        WaterSplatting \cite{li2025watersplatting}  & 22.97 & 99.78 \\
        SeaSplat \cite{yang2025seasplat}        & 85.00 & \underline{100.77} \\
        MarineSTD-GS \cite{marinestd_2025_acmmm}    & 72.31 & 41.13 \\
        Plenodium \cite{wu2025plenodium}       & \textbf{10.94} & 91.60 \\
        3D-UIR \cite{yuan2025_3duir}         & 48.00 & 42.74 \\
        \midrule
        3D-USE (Stage 1) & 8.73 & 143.71 \\
        3D-USE (Total)   & \underline{14.82} & \textbf{105.36} \\
        \bottomrule
    \end{tabular}
        \vspace{-5pt}
\end{table}

Qualitative results in Figure~\ref{fig:stage1_restlt} further show that explicit medium modeling improves background consistency and reduces medium-related artifacts. The depth visualizations confirm 3D-USE maintains coherent geometry and sharp object boundaries under underwater degradation.
Unlike methods that regress medium parameters using a global implicit field, MediumRBF shares information through adaptive local anchors, helping reduce coupling between object appearance and observer-dependent medium variations.

\begin{figure*}[!t]
	\centering
	\includegraphics[width=1\linewidth]{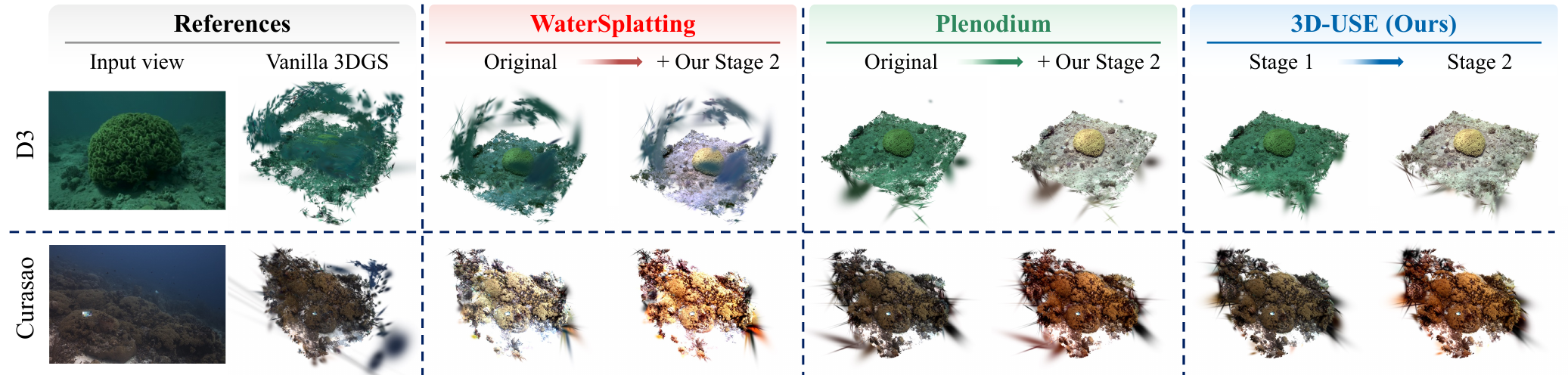}
    \vspace{-0.22in}
	\caption{Visualization of 3D scene representations before and after enhancement. Our Stage~1 reduces floating artifacts and medium-related color contamination, while Stage~2 improves scene appearance across different reconstruction models.}
	\label{fig:scene_visualization}
\end{figure*}

\begin{table*}[t]
	\caption{Impact of Stage~1 scene modeling on reconstruction and subsequent enhancement using the same Stage~2 procedure. Best and second-best results are bold and underlined.}
    \vspace{-5pt}
	\label{tab:stage1_ablation}
	\centering
	\small
	\begin{tabular*}{\textwidth}{@{\extracolsep{\fill}}lccccccc@{}}
		\toprule
		& \multicolumn{3}{c}{Scene Modeling Quality}
		& \multicolumn{4}{c}{Enhancement with Our Stage 2} \\
		\cmidrule(lr){2-4}\cmidrule(lr){5-8}
		Method & PSNR$\uparrow$ & SSIM$\uparrow$ & LPIPS$\downarrow$
		& UCIQE$\uparrow$ & URanker$\uparrow$ & MUSIQ$\uparrow$ & wLPIPS$\downarrow$ \\
		\midrule
		Vanilla 3DGS \cite{3DGS_tog_2023} & 27.559 & 0.868 & 0.225 & 0.5812 & 1.589 & 57.122 & 0.1824 \\
		WaterSplatting \cite{li2025watersplatting} & 28.976 & 0.878 & 0.180 & 0.5803 & 1.588 & 59.566 & 0.1572 \\
		Plenodium \cite{wu2025plenodium} & 29.850 & 0.885 & 0.172
		& 0.5810 & \underline{1.625} & 59.890 & 0.1558 \\
		MediumRBF w/o depth & \underline{30.292} & \underline{0.908} & \underline{0.154}
		& \underline{0.5814} & 1.624 & \textbf{60.169} & \underline{0.1550} \\
		\midrule
		Full Model (Ours) & \textbf{30.439} & \textbf{0.910} & \textbf{0.152}
		& \textbf{0.5816} & \textbf{1.630} & \underline{60.140} & \textbf{0.1540} \\
		\bottomrule
	\end{tabular*}
\end{table*}

\noindent\textbf{Enhancement Quality.}
Table~\ref{tab:enhancement} reports no-reference enhancement quality on held-out views. 3D-USE achieves the best URanker and MUSIQ on SeaThru-4, the best UCIQE and URanker on DRUVA-20, and leads all three metrics on D3 and D5. It also obtains the lowest average wLPIPS across 26 scenes, indicating the enhanced appearance remains stable across views rather than varying independently with each viewpoint.

Qualitative results in Figure~\ref{fig:stage2_restlt} show the limitations of physical restoration for scene enhancement. WaterSplatting~\cite{li2025watersplatting} and Plenodium~\cite{wu2025plenodium} exhibit clipped highlights and abrupt color changes, while SeaSplat~\cite{yang2025seasplat} and MarineSTD-GS~\cite{marinestd_2025_acmmm} retain residual haze and color casts. 3D-UIR~\cite{yuan2025_3duir} occasionally overcorrects foreground regions. In contrast, 3D-USE learns shared enhancement targets and jointly adjusts object and medium appearance, producing more balanced colors, clearer structures, and fewer spatial artifacts.

\noindent\textbf{Efficiency.}
Table~\ref{tab:efficiency} compares training time and rendering speed. Stage~2 increases per-scene training time from 8.73 to 14.82 minutes, with 3D-USE remaining the second fastest complete method after Plenodium. The full model renders at 105.36 FPS, the highest among all complete pipelines. These results show that scene-level enhancement incurs modest additional computation while preserving efficient real-time novel-view rendering.
Additional results and video demos are provided in the \textcolor{blue}{supplementary materials}.

\subsection{Ablation Study}
Our ablations examine how Stage~1 scene modeling affects scene reconstruction and enhancement, the complementary roles in Stage~2, and different strategies for incorporating 2D enhancement priors. All experiments use the four SeaThru scenes, D3, and D5, with results reported as scene-macro averages across the six scenes.

\noindent\textbf{Effect of Scene Modeling on Enhancement.} Table~\ref{tab:stage1_ablation} evaluates how different Stage~1 scene models affect reconstruction and subsequent enhancement under the same Stage~2 procedure. Compared with Vanilla 3DGS, explicit medium modeling substantially improves reconstruction fidelity and reduces cross-view differences after enhancement. Without depth supervision, MediumRBF improves reconstruction fidelity over Plenodium~\cite{wu2025plenodium}, with higher MUSIQ and lower wLPIPS after enhancement. Adding depth supervision further improves reconstruction and consistency, achieving the best PSNR, SSIM, LPIPS, and wLPIPS while maintaining competitive enhancement quality.

Figure~\ref{fig:scene_visualization} further shows how scene modeling affects enhancement. Vanilla 3DGS and WaterSplatting exhibit floating artifacts and fragmented structures, while Plenodium retains some localized artifacts despite its more coherent scene representation. Applying our Stage~2 to WaterSplatting and Plenodium adjusts their color and contrast, but cannot remove structural artifacts inherited from reconstruction, some of which become more noticeable after enhancement. In comparison, our Stage~1 produces a cleaner scene representation with fewer spurious Gaussians, providing more reliable geometry and visibility for associating enhancement cues across views. These comparisons show that physical restoration alone does not ensure satisfactory scene appearance, while appearance enhancement cannot repair errors already encoded in the 3D structure. Therefore, MediumRBF provides a more reliable basis for learning scene-level enhancement.

\begin{table}[t]
    \caption{Stage-2 ablation. Best and second-best results are bold and underlined.}
    \vspace{-5pt}
    \label{tab:stage2_ablation}
    \centering
    \small
    \setlength{\tabcolsep}{8pt}
    \begin{tabular}{@{}cccc@{}}
        \toprule
        Method & UCIQE $\uparrow$& URanker  $\uparrow$& MUSIQ $\uparrow$ \\
        \midrule
        Stage 1     & 0.481 & 0.431 & 55.740 \\
        Local only  & 0.521 & 1.071 & 52.176 \\
        Global only & \textbf{0.589} & \underline{1.490} & \underline{59.355} \\
        \midrule
        Full Model (Ours)    & \underline{0.582} & \textbf{1.630} & \textbf{60.140} \\
        \bottomrule
    \end{tabular}

\end{table}

\begin{figure}[!t]
	\centering
	\includegraphics[width=1\linewidth]{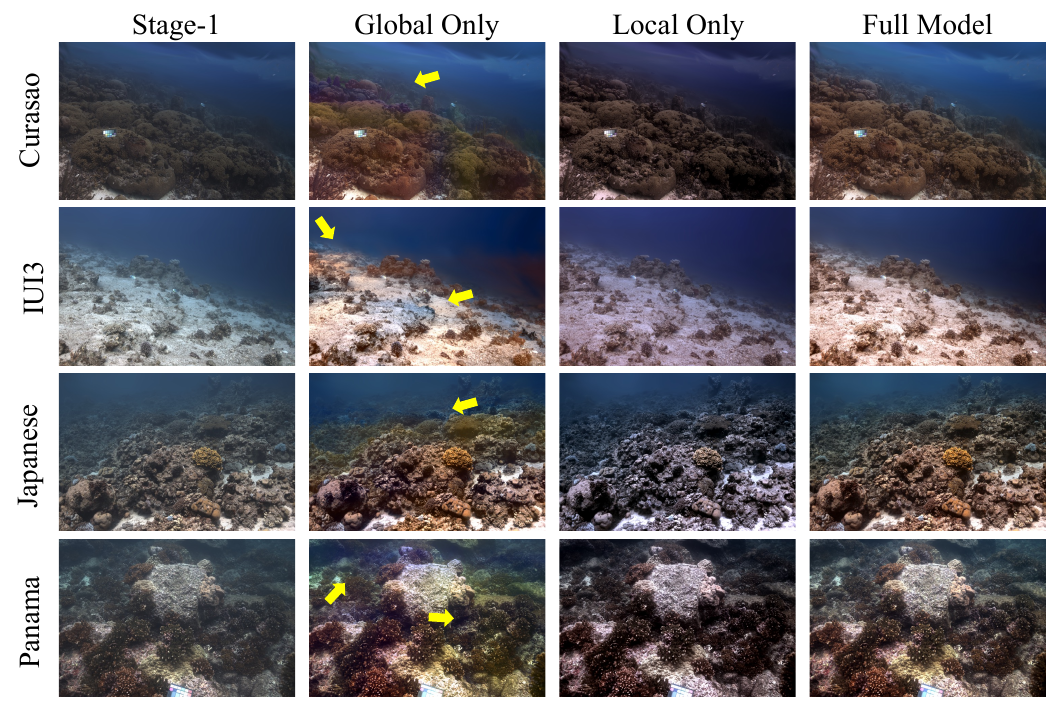}
    \vspace{-0.22in}
	\caption{Qualitative Stage-2 ablation results. Arrows highlight inconsistent enhancement in the global-only variant.%
    }
	\label{fig:stage2_ablation}
\end{figure}

\noindent\textbf{Complementary Enhancement Targets.} Table~\ref{tab:stage2_ablation} and Figure~\ref{fig:stage2_ablation} reveal complementary roles of the global and local transition targets. Global-only supervision produces the strongest scene-wide enhancement and achieves the highest UCIQE, but can introduce spatially uneven corrections, as reflected by the localized color discontinuities in Figure~\ref{fig:stage2_ablation}. In contrast, the zero-centered local targets constrain spatial variation but cannot establish the overall color and exposure change, leading to insufficient enhancement and a lower MUSIQ than the Stage~1 reference. Combining both targets suppresses these artifacts and achieves the best URanker and MUSIQ, with only a slight decrease in UCIQE relative to the global-only variant. This confirms that the global target determines the overall enhancement direction, while the local targets regularize its spatial realization.

\begin{figure}[!t]
	\centering
	\includegraphics[width=1\linewidth]{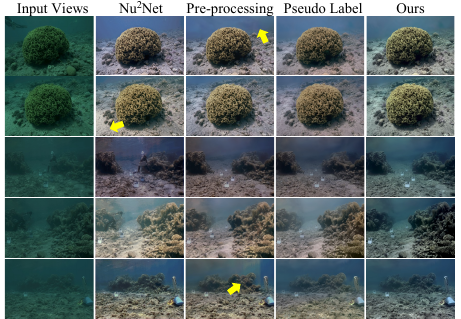}
    \vspace{-0.22in}
	\caption{Qualitative results of 2D UIE prior integration strategies. Arrows highlight view-dependent color shifts and local artifacts.
    }
	\label{fig:prior_integration}
\end{figure}

\noindent\textbf{2D Prior Integration Strategies.}
We examine whether directly using 2D enhancement results is sufficient for scene-level enhancement, and the effectiveness of ATC. Figure~\ref{fig:prior_integration} and Table~\ref{tab:prior_integration_quantitative} present comparison results, with all 2D enhancement models trained on UIEB. Direct Nu$^2$Net enhances each view independently and exhibits the largest cross-view differences. Pre-processing reconstructs the scene from enhanced training views, reducing wLPIPS compared with direct 2D enhancement but retaining local color artifacts despite achieving the highest MUSIQ.
RGB pseudo-label supervision retains the Stage~1 reconstruction and directly uses enhanced images as supervision targets. Although it achieves a low wLPIPS, Figure~\ref{fig:prior_integration} reveals residual color casts and softened local details. ATC instead calibrates view-wise enhancement proposals using a prior learned from paired 2D references and consolidates them into scene-global and Gaussian-local targets. Our method achieves higher UCIQE and URanker with nearly identical wLPIPS, albeit slightly lower MUSIQ. Overall, these results show that directly using 2D enhancement outputs, either for pre-processing or RGB supervision, does not reliably produce satisfactory scene-level enhancement. ATC instead calibrates and consolidates 2D appearance transitions into shared 3D targets, providing effective enhancement guidance without paired enhanced 3D references.

\begin{table}[t]
\caption{Quantitative results of 2D UIE prior integration strategies. Best and second-best results are bold and underlined.}
    \vspace{-5pt}
	\label{tab:prior_integration_quantitative}
	\centering
	\small
	\setlength{\tabcolsep}{3pt}
	\begin{tabular}{@{}lcccc@{}}
		\toprule
		Method
		& UCIQE$\uparrow$ & URanker$\uparrow$
		& MUSIQ$\uparrow$ & wLPIPS$\downarrow$ \\
		\midrule
		Nu$^2$Net
		& 0.567 & \underline{1.591}
		& 60.310 & 0.1969 \\
		Pre-processing
		& \underline{0.569} & 1.510
		& \textbf{61.120} & 0.1570 \\
		RGB Pseudo-Labels
		& 0.566 & 1.449
		& \underline{60.721} & \underline{0.1541} \\
		\midrule
		Ours
		& \textbf{0.582} & \textbf{1.630}
		& 60.140 & \textbf{0.1540} \\
		\bottomrule
	\end{tabular}
\end{table}

%% file: sec/5_conclusion.tex
\section{Conclusion}
We presented \textbf{3D-USE}, a two-stage framework for underwater scene-level enhancement without paired enhanced 3D supervision. It separates medium-aware reconstruction from appearance learning and consolidates appearance transitions learned from external paired 2D data into shared scene targets, rather than directly fitting independently enhanced images.
Experiments on real underwater scenes demonstrate improved perceptual quality and cross-view consistency while maintaining reconstruction fidelity.

%% file: sec/X_suppl.tex
\clearpage
\maketitlesupplementary

\appendix
This supplement presents the rendering preliminaries (Sec.~\ref{sec:supp_preliminaries}) and additional details of MediumRBF, ATC target construction, and U-BAF formulation and optimization (Secs.~\ref{sec:supp_reconstruction}--\ref{sec:supp_ubaf}). Additional experiments provide scene-wise reconstruction and enhancement results, qualitative comparisons, cross-view consistency evaluation, and computational cost (Sec.~\ref{sec:supp_evaluation}). Synchronized video comparisons illustrate visibility, local artifacts, and appearance consistency along matched camera trajectories (Sec.~\ref{sec:supp_video}). We also discuss the limitations of per-scene optimization and future directions toward feed-forward reconstruction and enhancement (Sec.~\ref{sec:supp_limitations}).

\section{Preliminaries}
\label{sec:supp_preliminaries}
\subsection{3D Gaussian Splatting}
3D Gaussian Splatting (3DGS)~\cite{3DGS_tog_2023} represents a scene with Gaussian primitives, each defined by a center $\mu$, covariance $\Sigma$, opacity $\alpha$, and view-dependent color. A rotation matrix $R$ and scaling matrix $S$ parameterize the covariance as
	\begin{equation}
		\Sigma = RSS^\top R^\top,
	\end{equation}
For rasterization, the covariance is projected onto the image plane. Let $W$ denote the linear part of the world-to-camera transformation and $J$ the Jacobian of the local projective approximation. The projected covariance is
	\begin{equation}
		\Sigma' = JW\Sigma W^\top J^\top.
	\end{equation}
For pixel $p$, the contributing Gaussians are ordered by depth in $\mathcal G_p$, and their colors are composited as
	\begin{equation}
		\mathbf C = \sum_{i \in \mathcal G_p} \mathbf c_i \alpha_i \prod_{j=1}^{i-1}(1-\alpha_j),
	\end{equation}
where $\mathbf c_i$ and $\alpha_i$ denote the color and projected opacity of Gaussian $i$ at the pixel.

\subsection{Underwater Gaussian Compositor}
\label{sec:supp_compositor}
The underwater compositor combines Gaussian radiance with light attenuation and backscatter~\cite{li2025watersplatting,wu2025plenodium}. MediumRBF provides effective medium parameters for each ray, which are held constant along that ray. The rendered color consists of the attenuated object contribution and the accumulated medium contribution.

For a ray $\mathbf r(s)=\mathbf o+s\boldsymbol\omega$ with unit direction $\boldsymbol\omega$, let $\{s_i\}_{i=1}^{N}$ be the depth-ordered Gaussian distances, with $s_0=0$. Foreground occlusion is described by the object transmittance
	\begin{equation}
		T_i^{\mathrm{obj}}=\prod_{j=1}^{i-1}(1-\alpha_j),
	\end{equation}
where $\alpha_j$ is the projected opacity of Gaussian $j$. The object contribution is attenuated according to its distance from the camera:
	\begin{equation}
		\widehat{\mathbf C}^{\mathrm{obj}}
		=
		\sum_{i=1}^{N}
		T_i^{\mathrm{obj}}\alpha_i\,
		\mathbf c_i\odot
		\exp(-\boldsymbol\beta_{\mathrm{attn}}s_i),
	\end{equation}
	where $\boldsymbol\beta_{\mathrm{attn}}$ is the RGB attenuation coefficient along the ray and $\odot$ denotes element-wise multiplication.

Let $\mathbf c_m$ and $\boldsymbol\beta_{\mathrm{bs}}$ denote the medium color and RGB backscatter coefficient. Backscatter accumulated between successive Gaussians contributes
	\begin{equation}
		\begin{aligned}
            \widehat{\mathbf C}^{\mathrm{med}}={}\sum_{i=1}^{N}T_i^{\mathrm{obj}}\,\mathbf c_m\odot
            \left[\exp(-\boldsymbol\beta_{\mathrm{bs}}s_{i-1})-\exp(-\boldsymbol\beta_{\mathrm{bs}}s_i)\right].
        \end{aligned}
	\end{equation}
Beyond the last Gaussian, the remaining medium contribution is weighted by $T_{N+1}^{\mathrm{obj}}=\prod_{j=1}^{N}(1-\alpha_j)$:
	\begin{equation}
		\widehat{\mathbf C}^{\mathrm{med}}_{\infty}
		=
		T_{N+1}^{\mathrm{obj}}\mathbf c_m\odot
		\exp(-\boldsymbol\beta_{\mathrm{bs}}s_N).
	\end{equation}
The complete rendering combines these contributions:
	\begin{equation}
		\widehat{\mathbf C}
		=\widehat{\mathbf C}^{\mathrm{obj}}
		+\widehat{\mathbf C}^{\mathrm{med}}
		+\widehat{\mathbf C}^{\mathrm{med}}_{\infty}.
	\end{equation}
Exponentials are evaluated channel-wise, and $T_i^{\mathrm{obj}}$ accounts only for occlusion by the Gaussians.

\section{Medium-Aware Reconstruction}
\label{sec:supp_reconstruction}
\paragraph{Medium Representation.} MediumRBF estimates observer-dependent water effects through shared local anchors~\cite{NeuRBF_2023_ICCV}. Their SH coefficients are aggregated at the observer position and decoded along the viewing direction, coupling local spatial support with directional variation. The output mapping $\Phi$ enforces nonnegative medium color and applies Softplus to attenuation and backscatter.

\paragraph{Reconstruction Objective.} To emphasize reconstruction errors in dark regions, we use inverse-intensity weighting~\cite{wu2025plenodium}. Let $\widehat{\mathbf I}$ and $\mathbf I^u$ denote the rendered and observed underwater images. The weights are defined from the detached rendering as
	\begin{equation}
		\mathbf W
		=
		\left(
		\operatorname{sg}(\widehat{\mathbf I})+\epsilon
		\right)^{-1}.
		\label{eq:supp_reg_weight}
	\end{equation}
Here, $\operatorname{sg}(\cdot)$ denotes stop-gradient, and $\epsilon>0$ denotes a small numerical stabilizer throughout this supplement. The weighted photometric and structural losses are
	\begin{equation}
		\begin{aligned}
			\mathcal L_{\mathrm{Reg\text{-}L1}}
			&=
			\left\|
			\mathbf W\odot
			(\widehat{\mathbf I}-\mathbf I^{u})
			\right\|_1,\\
			\mathcal L_{\mathrm{Reg\text{-}MS\text{-}SSIM}}
			&=
			\mathcal L_{\mathrm{MS\text{-}SSIM}}
			\left(
			\mathbf W\odot\mathbf I^{u},
			\mathbf W\odot\widehat{\mathbf I}
			\right).
		\end{aligned}
		\label{eq:supp_reg_reconstruction}
	\end{equation}
These terms form the reconstruction objective
	\begin{equation}
		\mathcal L_{\mathrm{rec}}
		=
		(1-\lambda)\mathcal L_{\mathrm{Reg\text{-}L1}}
		+
		\lambda\mathcal L_{\mathrm{Reg\text{-}MS\text{-}SSIM}}.
		\label{eq:supp_reconstruction}
	\end{equation}
Relative-depth supervision complements the photometric objective. We align the shifted inverse-depth representation $\widehat{\mathbf d}$ of rendered depth $\widehat{\mathbf D}$ with the Depth Anything V2 prior $\mathbf d^{\mathrm{DA2}}$~\cite{Depth_Anything_V2} using a correlation loss~\cite{marinestd_2025_acmmm}:
	\begin{equation}
		\mathcal L_{\mathrm{depth}}
		=1-\operatorname{Pearson}
		(\widehat{\mathbf d},\mathbf d^{\mathrm{DA2}}).
		\label{eq:supp_depth}
	\end{equation}
The correlation is evaluated over valid pixels and accommodates the unknown scale and offset of the monocular disparity prior.

\paragraph{Training Details.} MediumRBF uses eight anchors, top-$K$ aggregation with $K=4$, and degree-3 SH. Stage~1 is optimized for 15k iterations, with Gaussian densification ending at 10k. The reconstruction mixing weight and depth-loss weight are $\lambda=0.2$ and $\lambda_d=0.1$, respectively.

\section{Appearance Transition Consensus}
\label{sec:supp_atc}
\paragraph{Transition Extraction and Calibration.} The extractor $\mathcal E$ represents an appearance change as a global color transform followed by spatially varying residual corrections. For an RGB color $\mathbf c$, the global affine is parameterized relative to the identity:
\begin{equation}
    \mathcal T^g(\mathbf c)=(\boldsymbol\Lambda_3+\Delta\mathbf A)\mathbf c+\Delta\mathbf b.
    \label{eq:supp_global_operator}
\end{equation}
Here, $\boldsymbol\Lambda_3$ is the $3\times3$ identity matrix. The parameter vector $\mathbf t^g$ stacks the rows of $[\Delta\mathbf A\mid\Delta\mathbf b]$. Within each local cell $p$, gain residuals $\mathbf d_p$ and biases $\mathbf b_p$ describe the remaining adjustment between corresponding source and target colors $(\mathbf c_a,\mathbf c_b)$:
\begin{equation}
    \mathbf c_b\approx(\mathbf 1+\mathbf d_p)\odot\mathcal T^g(\mathbf c_a)+\mathbf b_p.
    \label{eq:supp_local_operator}
\end{equation}
The local transition vector $\boldsymbol\Delta_p^\ell=[\mathbf d_p^\top,\mathbf b_p^\top]^\top$ is zero for an identity correction. Both operators are estimated by ridge regression, which remains differentiable with respect to the target image.

The calibrator learns to correct UIE proposals using 800 underwater--enhanced image pairs from UIEB~\cite{UIEB_TIP_2019}, excluding all target scenes. For each pair $(\mathbf I^u,\mathbf I^e)$, the reference and proposed transitions are
\begin{equation}
    \begin{aligned}
        (\mathbf t^{g,*},\boldsymbol\Delta^{\ell,*})&=\mathcal E(\mathbf I^u,\mathbf I^e),\\
        (\mathbf t^{g,n},\boldsymbol\Delta^{\ell,n})&=\mathcal E(\mathbf I^u,\mathcal N(\mathbf I^u)),
    \end{aligned}
    \label{eq:supp_pair_operators}
\end{equation}
where $\mathcal N$ is the frozen Nu$^2$Net from URanker~\cite{uranker_aaai_2023}. The residual calibrator is conditioned on source-image statistics, with its local head additionally receiving patch statistics and the global prediction. It is trained with scale-normalized Smooth-L1 losses~\cite{huber1964robust}, weighted locally by patch-fitting confidence, and remains fixed during scene optimization.

\paragraph{Scene-Global Consensus.} For each captured view $v$, we evaluate the calibrated candidate $\widehat{\mathcal T}_v^g$ on a common RGB probe set $\mathbf P$, yielding $\mathbf Y_v=\widehat{\mathcal T}_v^g(\mathbf P)$. The scene-global transform is the affine fit to the geometric median of these responses:
\begin{equation}
    \begin{aligned}
        \mathbf Y_{\mathrm{sc}}&=\arg\min_{\mathbf Y}\sum_v\|\mathbf Y-\mathbf Y_v\|_F,\\
        \overline{\mathcal T}_{\mathrm{sc}}^g&=\arg\min_{\mathcal T}\|\mathcal T(\mathbf P)-\mathbf Y_{\mathrm{sc}}\|_F^2.
    \end{aligned}
    \label{eq:supp_global_consensus}
\end{equation}

\paragraph{Gaussian-Local Consensus.} Local enhancement cues are consolidated on the fixed Gaussian geometry. Conditioned on $\overline{\mathcal T}_{\mathrm{sc}}^g$, the local head predicts view-wise correction maps. Spatial centering and bilinear resampling yield the pixel-wise residual $\widehat{\boldsymbol\Delta}_v^\ell$. Let $w_{iv}(p)$ denote the splatting weight of Gaussian $i$ at pixel $p$ in view $v$. Its visibility support and lifted residual are
\begin{equation}
	\begin{aligned}
		\omega_{iv}&=\sum_p w_{iv}(p),\\
		\widehat{\boldsymbol\Delta}_{iv}^{\ell}
		&=\frac{\sum_p w_{iv}(p)\,\widehat{\boldsymbol\Delta}_v^\ell(p)}
		{\omega_{iv}+\epsilon}.
	\end{aligned}
	\label{eq:supp_local_lifting}
\end{equation}
The weights account for the projected footprint, opacity, and occlusion of each Gaussian, excluding medium attenuation. Only observations with $\omega_{iv}>\epsilon$ contribute to fusion.

Visibility-weighted fusion with Cauchy reweighting in scale-normalized units gives a fixed target $\overline{\boldsymbol\Delta}_i^\ell$ for each Gaussian. Local supervision is restricted to Gaussians observed in at least two views. Their targets and binary support indicators are splatted into each view, and normalization by the accumulated support yields $\overline{\boldsymbol\Delta}_v^\ell$. The same projected support provides the local loss weights.

\section{Underwater Bilateral Appearance Field}
\label{sec:supp_ubaf}
\paragraph{Bilateral Field.} U-BAF parameterizes local color corrections jointly by spatial location and base appearance~\cite{gharbi2017deepbilateral,BilRF_tog_2024}. The scalar guide $\gamma(\bar{\mathbf c}_i)$ is learned from the fixed, view-independent Stage~1 base color $\bar{\mathbf c}_i$. Four one-dimensional feature grids encode the three spatial coordinates and the guide. Their interpolated features are multiplied element-wise and decoded into RGB log-gains and biases. Subtracting the mean over Gaussians yields the centered corrections $\mathbf a_i^\ell$ and $\mathbf b_i^\ell$, reducing overlap with the scene-global transform.

\paragraph{Object and Medium Appearance.} The scene-global affine $\mathcal T_\theta^g(\mathbf c)=\mathbf A_\theta\mathbf c+\mathbf b_\theta$ sets the shared color transformation, while the local field provides spatially and appearance-dependent corrections. The enhanced Gaussian radiance is
\begin{equation}
    \mathbf c_i^+(\boldsymbol\omega)=\exp(\mathbf a_i^\ell)\odot\mathcal T_\theta^g\!\left(\mathbf c_i(\boldsymbol\omega)\right)+\mathbf b_i^\ell.
    \label{eq:supp_enhanced_radiance}
\end{equation}
The same global affine and appearance field transform medium color, with the field evaluated at the camera position. Its local log-gains and biases are centered over a uniform guide grid. For each ray, gains and biases are interpolated at $\gamma(\mathbf c_m)$ after exponentiating the centered log-gains, and applied to the globally transformed medium color. Attenuation and backscatter are adjusted by learned, bounded positive scalars shared across the scene:
\begin{equation}
    \boldsymbol\beta_j^+=r_j\boldsymbol\beta_j,\qquad j\in\{\mathrm{attn},\mathrm{bs}\}.
    \label{eq:supp_medium_scaling}
\end{equation}
Each scalar preserves the RGB channel proportions of its Stage~1 coefficient. The underwater compositor combines the transformed Gaussian radiance and medium parameters to produce $\mathbf I_v^+$.

\paragraph{Enhancement Objective.} Supervision acts on the appearance change realized by the complete underwater rendering, accounting jointly for object and medium adjustments. Applying the fixed extractor to the Stage~1 rendering $\mathbf I_v^{\mathrm{rec}}$ and enhanced rendering $\mathbf I_v^+$ gives
\begin{equation}
    (\mathbf t_v^g,\boldsymbol\Delta_v^\ell)=\mathcal E\!\left(\operatorname{sg}(\mathbf I_v^{\mathrm{rec}}),\mathbf I_v^+\right),
    \label{eq:supp_realized_transition}
\end{equation}
The local map $\boldsymbol\Delta_v^\ell$ is spatially centered to give $\widetilde{\boldsymbol\Delta}_v^\ell$, which is compared with the projected ATC target on the local grid $\mathcal P_v$. Resampling the target and its support to this grid yields $\overline{\boldsymbol\Delta}_{v,p}^\ell$ and weights $W_{vp}\in[0,1]$.

Let $\overline{\mathbf t}_{\mathrm{sc}}^g$ be the identity-centered parameter vector of the fixed global target. To balance transition components with different scales, we normalize their errors by fixed vectors $\mathbf s_g$ and $\mathbf s_\ell$. These contain the component-wise interquartile ranges of reference-minus-proposal parameters in the calibration data, lower-bounded by $\epsilon$:
\begin{equation}
    \begin{aligned}
        \mathbf e_v^g&=\frac{\mathbf t_v^g-\overline{\mathbf t}_{\mathrm{sc}}^g}{\mathbf s_g},\\
        \mathbf e_{vp}^\ell&=\frac{\widetilde{\boldsymbol\Delta}_{v,p}^\ell-\overline{\boldsymbol\Delta}_{v,p}^\ell}{\mathbf s_\ell},
    \end{aligned}
    \label{eq:supp_normalized_errors}
\end{equation}
Division is component-wise. The global loss averages the Smooth-L1 penalty across components, while the local loss additionally weights each cell by its projected support:
\begin{equation}
    \begin{aligned}
        \mathcal L_{\mathrm{global}}&=\operatorname*{mean}_q\,\phi_{\mathrm{SL1}}(e_{v,q}^g),\\
        \mathcal L_{\mathrm{local}}&=\frac{\sum_{p\in\mathcal P_v}W_{vp}\operatorname*{mean}_q\,\phi_{\mathrm{SL1}}(e_{vp,q}^\ell)}{\max\!\left(\sum_{p\in\mathcal P_v}W_{vp},\epsilon\right)}.
    \end{aligned}
    \label{eq:supp_stage2_terms}
\end{equation}
Here, $p$ and $q$ index local cells and transition components, and $\phi_{\mathrm{SL1}}$ is the unit-threshold Smooth-L1 penalty. These reductions specify the $\|\cdot\|_{\mathrm{S1}}$ notation in the main paper. The full objective is
\begin{equation}
    \mathcal L_{\mathrm{S2}}=\lambda_g\mathcal L_{\mathrm{global}}+\lambda_\ell\mathcal L_{\mathrm{local}}+\lambda_{\mathrm{tv}}\mathcal L_{\mathrm{TV}},
    \label{eq:supp_stage2_objective}
\end{equation}
The total-variation term $\mathcal L_{\mathrm{TV}}$ averages absolute first differences along the factor grids. Optimization updates the appearance field, global affine, and medium scalars, while keeping the Stage~1 reconstruction and ATC targets fixed.

\begin{figure*}[!th]
	\centering
	\includegraphics[width=1\textwidth]{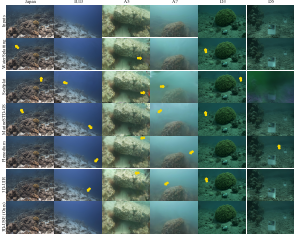}
	\caption{\textbf{Additional qualitative comparison of underwater novel-view reconstruction.}  Columns correspond to Japan (Japanese Gardens), IUI3, A5, A7, D3, and D5. The first row shows the held-out underwater views, followed by the reconstruction results of WaterSplatting, SeaSplat, MarineSTD-GS, Plenodium, 3D-UIR, and 3D-USE.}
	\label{fig:supp_reconstruction_comparison}
\end{figure*}

\begin{table*}[t]
	\caption{Scene-wise reconstruction results on SeaThru-4 held-out views.
		Best and second-best results are bold and underlined.}
	\label{tab:supp_seathru_reconstruction}
	\centering
	\small
	\begin{tabular*}{\textwidth}{@{\extracolsep{\fill}}l*{12}{c}@{}}
		\toprule
		& \multicolumn{3}{c}{IUI3}
		& \multicolumn{3}{c}{Curasao}
		& \multicolumn{3}{c}{Japanese Gardens}
		& \multicolumn{3}{c}{Panama} \\
		\cmidrule(lr){2-4}\cmidrule(lr){5-7}
		\cmidrule(lr){8-10}\cmidrule(lr){11-13}
		Method
		& PSNR & SSIM & LPIPS
		& PSNR & SSIM & LPIPS
		& PSNR & SSIM & LPIPS
		& PSNR & SSIM & LPIPS \\
		\midrule
		WaterSplatting
		& 29.330 & 0.891 & 0.206
		& 31.701 & 0.951 & 0.114
		& 24.732 & 0.892 & 0.120
		& 31.598 & 0.941 & 0.076 \\
		SeaSplat
		& 27.552 & 0.853 & 0.209
		& 29.323 & 0.892 & 0.191
		& 22.777 & 0.867 & 0.190
		& 28.932 & 0.899 & 0.146 \\
		MarineSTD-GS
		& \textbf{30.433} & 0.890 & \textbf{0.202}
		& 33.688 & 0.939 & 0.143
		& \textbf{26.337} & 0.881 & 0.152
		& 29.678 & 0.893 & 0.145 \\
		Plenodium
		& 30.179 & \underline{0.892} & \underline{0.205}
		& \underline{33.861} & \underline{0.955} & \underline{0.105}
		& 24.872 & \underline{0.895} & \textbf{0.115}
		& \textbf{32.413} & \underline{0.945} & \underline{0.075} \\
		3D-UIR
		& 28.640 & 0.868 & 0.233
		& 30.722 & 0.896 & 0.178
		& 23.592 & 0.884 & 0.162
		& 30.085 & 0.910 & 0.142 \\
		\midrule
		\textbf{3D-USE (Ours)}
		& \underline{30.215} & \textbf{0.896} & 0.206
		& \textbf{33.965} & \textbf{0.957} & \textbf{0.104}
		& \underline{24.957} & \textbf{0.896} & \underline{0.116}
		& \underline{32.337} & \textbf{0.947} & \textbf{0.071} \\
		\bottomrule
	\end{tabular*}

\end{table*}

\paragraph{Training Details.} U-BAF uses factor grids of resolution 16 and rank $R=8$, with a two-layer decoder of width 64. Starting from the fixed Stage~1 scene, we optimize the appearance transformation for 5k iterations using Adam, with learning rates of $10^{-3}$ for the appearance field and $2\times10^{-4}$ for the medium ratios. We set $(\lambda_g,\lambda_\ell,\lambda_{\mathrm{tv}})=(1,1,10^{-4})$.

\begin{figure*}[!t]
	\centering
	\includegraphics[width=1\textwidth]{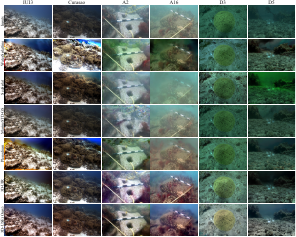}
	\caption{\textbf{Additional qualitative comparison of enhanced novel-view renderings.}
		Columns correspond to IUI3, Curasao, A2, A16, D3, and D5.
		Rows show the corresponding underwater inputs and the results of
		WaterSplatting, SeaSplat, MarineSTD-GS, Plenodium, 3D-UIR, and 3D-USE.}
	\label{fig:supp_enhancement_comparison}
\end{figure*}

\begin{table*}[t]
	\caption{Scene-wise no-reference enhancement quality on SeaThru-4 held-out views. Higher is better for all metrics. Best and second-best results are bold and underlined.}
	\label{tab:supp_seathru_enhancement}
	\centering
	\small
	\setlength{\tabcolsep}{1mm}
	\begin{tabular}{@{}l*{12}{c}@{}}
		\toprule
		& \multicolumn{3}{c}{IUI3}
		& \multicolumn{3}{c}{Curasao}
		& \multicolumn{3}{c}{Japanese Gardens}
		& \multicolumn{3}{c}{Panama} \\
		\cmidrule(lr){2-4}\cmidrule(lr){5-7}
		\cmidrule(lr){8-10}\cmidrule(lr){11-13}
		Method
		& UCIQE & URanker & MUSIQ
		& UCIQE & URanker & MUSIQ
		& UCIQE & URanker & MUSIQ
		& UCIQE & URanker & MUSIQ \\
		\midrule
		WaterSplatting
		& 0.509 & 1.045 & 57.549
		& 0.547 & \textbf{1.924} & \textbf{66.755}
		& 0.497 & 1.555 & 56.981
		& 0.480 & 0.920 & 52.831 \\
		SeaSplat
		& 0.579 & 1.503 & 51.292
		& \underline{0.586} & 1.247 & 60.171
		& 0.601 & 1.922 & 53.844
		& 0.586 & 1.549 & 52.572 \\
		Plenodium
		& 0.511 & 0.826 & \underline{60.681}
		& 0.499 & 0.831 & 61.631
		& 0.507 & 1.489 & \underline{57.784}
		& 0.496 & 1.086 & 53.384 \\
		MarineSTD-GS
		& 0.546 & 1.161 & 58.370
		& 0.477 & 0.087 & 58.286
		& 0.543 & 1.329 & 55.681
		& 0.520 & 1.094 & 51.222 \\
		3D-UIR
		& \underline{0.597} & \underline{1.558} & 49.705
		& \textbf{0.593} & \underline{1.258} & 62.398
		& \underline{0.621} & \underline{1.925} & 52.173
		& \textbf{0.623} & \textbf{1.890} & \underline{53.886} \\
		\midrule
		\textbf{3D-USE (Ours)}
		& \textbf{0.629} & \textbf{1.733} & \textbf{61.373}
		& 0.564 & 1.000 & \underline{64.469}
		& \textbf{0.632} & \textbf{2.112} & \textbf{59.718}
		& \underline{0.608} & \underline{1.873} & \textbf{53.964} \\
		\bottomrule
	\end{tabular}

\end{table*}

\section{Additional Experiments}
\label{sec:supp_evaluation}

\subsection{Evaluation Protocol}

\paragraph{Image-quality Metrics.} Reconstruction metrics are computed on held-out views in the display range $[0,1]$. SSIM is computed on RGB images, and LPIPS uses the AlexNet backbone. We first average image scores within a scene and then report the unweighted mean of scene scores for SeaThru-4 and DRUVA-20.

Enhancement quality is evaluated on held-out test views. We compute UCIQE~\cite{uciqe_TIP_2015} using the same reference implementation for every method. URanker and MUSIQ are computed with their official pretrained models~\cite{uranker_aaai_2023, MUSIQ_iccv_2021}; inputs whose long side exceeds 1024 pixels are resized while preserving aspect ratio. For all three metrics, view scores are averaged within each scene and dataset values are unweighted scene-macro means. Thus, large DRUVA scenes do not dominate scenes with fewer held-out cameras.
\paragraph{Cross-view Consistency.} Following the warped LPIPS (wLPIPS) protocol used by StylizedNeRF~\cite{stylizednerf_cvpr_2022}, we evaluate cross-view consistency using LPIPS between adjacent views after optical-flow warping. RAFT-Large~\cite{RAFT_eccv_2020} estimates forward and backward flow from the raw inputs. All methods share the same flow and forward--backward validity masks. Pair scores are averaged within each scene and then across scenes.

\subsection{Quantitative Comparisons}

\paragraph{Scene-wise Results.} Tables~\ref{tab:supp_seathru_reconstruction} and~\ref{tab:supp_seathru_enhancement} decompose the SeaThru-4 macro averages reported in the main paper into individual scenes. 3D-USE achieves the best SSIM in all four scenes and the best or second-best LPIPS in Curasao, Japanese Gardens, and Panama. For enhancement, it leads all three metrics on IUI3 and Japanese Gardens, achieves the best MUSIQ and second-best UCIQE and URanker on Panama, and ranks second in MUSIQ on Curasao. The gains thus extend across multiple scenes and water conditions.

\paragraph{Cross-view Consistency.} Table~\ref{tab:supp_warped_lpips} shows that 3D-USE obtains the lowest wLPIPS on both multi-scene benchmarks and the lowest 26-scene macro average reported in the main paper. The single-scene D3 and D5 results are less favorable, so this metric is used as complementary evidence of overall cross-view stability rather than a claim of uniform dominance on every trajectory.

\begin{table}[t]
    \caption{Dataset-wise cross-view consistency measured by wLPIPS
			(lower is better). Best and second-best results are bold and
			underlined.}
		\label{tab:supp_warped_lpips}
		\centering
		\small
		\setlength{\tabcolsep}{4pt}
		\begin{tabular}{@{}lcccc@{}}
			\toprule
			& \multicolumn{4}{c}{wLPIPS$\downarrow$} \\
			\cmidrule(lr){2-5}
			Method & SeaThru-4 & DRUVA-20 & D3 & D5 \\
			\midrule
			WaterSplatting & 0.1524 & 0.1593 & 0.2186 & 0.2477 \\
			SeaSplat       & 0.1465 & 0.2355 & 0.2110 & 0.2428 \\
			Plenodium      & \underline{0.1389} & \underline{0.1419}
			& \underline{0.2096} & 0.2456 \\
			MarineSTD-GS   & 0.1446 & 0.2222 & \textbf{0.2082}
			& 0.2308 \\
			3D-UIR         & 0.1498 & 0.2392 & 0.2267 & \textbf{0.1834} \\
			\midrule
			\textbf{3D-USE (Ours)}
			& \textbf{0.1259} & \textbf{0.1333}
			& 0.2196 & \underline{0.2006} \\
			\bottomrule
		\end{tabular}

	\end{table}

\subsection{Qualitative Comparisons}

\paragraph{Reconstruction Quality.} Figure~\ref{fig:supp_reconstruction_comparison} presents additional held-out reconstruction results across six scenes with varying contents and water conditions. On Japan and IUI3, differences are mainly visible in fine coral structures and distant backgrounds. The A5 and A7 scenes further reveal medium-related artifacts around object boundaries and hazy regions. WaterSplatting and SeaSplat exhibit local blurring or color deviations, with SeaSplat showing a severe chromatic failure on D5. MarineSTD-GS and Plenodium retain localized opacity and background artifacts, while 3D-UIR shows residual appearance inconsistencies in some regions. In contrast, 3D-USE more faithfully reproduces the underwater observations, preserving object boundaries, fine structures, and background appearance with fewer local artifacts. These results highlight the benefit of medium-aware modeling for accurate reconstruction across diverse underwater conditions.

\paragraph{Enhancement Quality.} Figure~\ref{fig:supp_enhancement_comparison} extends the enhancement comparison to six scenes with different contents and dominant color casts. The baselines exhibit a recurring trade-off between enhancement strength and spatial stability. WaterSplatting strongly increases contrast but clips bright seabed regions in IUI3 and Curasao and darkens distant regions in A16 and D5. SeaSplat applies milder correction yet retains prominent green casts in A2, A16, and D5. MarineSTD-GS often remains close to the degraded input and shows localized artifacts in A2, whereas Plenodium produces aggressive color and contrast changes in IUI3. The 3D-UIR results vary more markedly across scenes, including magenta-biased corrections in A2 and A16 and limited recovery in D5. In comparison, 3D-USE maintains a more consistent scene-wide grade while preserving local structures such as coral texture, ropes, and calibration targets. Its behavior across blue-, green-, and haze-dominated scenes is consistent with the intended roles of the shared transition consensus and Gaussian-local appearance field: the former determines the global enhancement direction, while the latter accommodates spatially varying corrections without independently processing each rendered view.

\subsection{Computational Cost}
On SeaThru-4, 3D-USE takes 8.73 minutes for Stage~1 and an additional 6.09 minutes for Stage~2, totaling 14.82 minutes per scene on a single RTX~3090. Timing covers the training loops and scene-specific ATC target construction; offline preprocessing, shared calibrator training, rendering, and evaluation are excluded. At $1400\times933$ resolution, the Stage~1 and enhanced scenes render at 143.71 and 105.36 FPS, respectively.

\section{Video Supplement}
\label{sec:supp_video}
The accompanying video supplement compares reconstruction and scene-level enhancement along matched camera trajectories, enabling inspection of visibility, local artifacts, and cross-view appearance consistency. Extract the archive and open \texttt{index.html} locally to select scenes and methods, view synchronized side-by-side or wipe comparisons, and inspect enlarged details in side-by-side mode. For enhancement, our Stage~1 rendering provides the default unenhanced reference.

\section{Limitations and Future Work}
\label{sec:supp_limitations}
3D-USE inherits the per-scene optimization paradigm of 3DGS-based reconstruction, requiring separate reconstruction and enhancement optimization for each new scene. This introduces additional computational overhead and limits the efficiency of scene construction. A promising direction is feed-forward underwater scene reconstruction and enhancement, where a generalizable model predicts a medium-aware 3D representation and its enhanced appearance directly from multi-view observations while preserving cross-view consistency.

%% file: main.bib
@ARTICLE{yuan_JOE_2025,
	author={Yuan, Jieyu and Zhang, Yuhan and Cai, Zhanchuan},
	journal={IEEE JOE}, 
	title={Underwater Scene Enhancement via Adaptive Color Analysis and Multispace Fusion}, 
	year={2025},
	volume={50},
	number={4},
	pages={2516-2528},
	doi={10.1109/JOE.2025.3591405}}

@INPROCEEDINGS{Huang_2023_CVPR,
  author={Huang, Shirui and Wang, Keyan and Liu, Huan and Chen, Jun and Li, Yunsong},
  booktitle={CVPR}, 
  title={Contrastive Semi-Supervised Learning for Underwater Image Restoration via Reliable Bank}, 
  year={2023},
  volume={},
  number={},
  pages={18145-18155},
  doi={10.1109/CVPR52729.2023.01740}}

@INPROCEEDINGS{DRUVA_ICCV_2023,
	author={Varghese, Nisha and Kumar, Ashish and Rajagopalan, A. N.},
	booktitle={ICCV}, 
	title={Self-supervised Monocular Underwater Depth Recovery, Image Restoration, and a Real-sea Video Dataset}, 
	year={2023},
	volume={},
	number={},
	pages={12248-12258},
	doi={10.1109/ICCV51070.2023.01125}}

@inproceedings{boittiaux2024sucre,
	author={Boittiaux, Cl\'ementin and Marxer, Ricard and Dune, Claire and Arnaubec, Aur\'elien and Ferrera, Maxime and Hugel, Vincent},
	title={{SUCRe}: Leveraging Scene Structure for Underwater Color Restoration},
	booktitle={3DV},
	year={2024}
}

@inproceedings{xie2024uveb,
  title={{UVEB}: A Large-scale Benchmark and Baseline Towards Real-World Underwater Video Enhancement},
  author={Xie, Yaofeng and Kong, Lingwei and Chen, Kai and Zheng, Ziqiang and Yu, Xiao and Yu, Zhibin and Zheng, Bing},
  booktitle={CVPR},
  pages={22358--22367},
  year={2024}
}

@ARTICLE{PUGAN_TIP_2023,
	author={Cong, Runmin and Yang, Wenyu and Zhang, Wei and Li, Chongyi and Guo, Chun-Le and Huang, Qingming and Kwong, Sam},
	journal={IEEE TIP}, 
	title={{PUGAN}: Physical Model-Guided Underwater Image Enhancement Using GAN With Dual-Discriminators}, 
	year={2023},
	volume={32},
	number={},
	pages={4472-4485},
	doi={10.1109/TIP.2023.3286263}}

@ARTICLE{W2WDiff_tgrs_2026,
  author={Zhang, Yuanlin and Yuan, Jieyu and Chen, Xiao and Tang, Xiongxin and Chen, Qiao and Wang, Yiquan and Li, Chongyi},
  journal={IEEE TGRS}, 
  title={{W2WDiff}: Generalizing Underwater Diffusion Model via Unsupervised Underwater Conversion}, 
  year={2026},
  volume={64},
  number={},
  pages={1-13},
  doi={10.1109/TGRS.2025.3629979}}

@article{3DGS_tog_2023,
  title   = {3D Gaussian Splatting for Real-Time Radiance Field Rendering},
  author  = {Kerbl, Bernhard and Kopanas, Georgios and Leimk{\"u}hler, Thomas and Drettakis, George},
  journal = {ACM TOG},
  volume  = {42},
  number  = {4},
  year    = {2023},
  doi     = {10.1145/3592433}
}

@inproceedings{martinbrualla2021nerfw,
  title     = {NeRF in the Wild: Neural Radiance Fields for Unconstrained Photo Collections},
  author    = {Martin-Brualla, Ricardo and Radwan, Noha and Sajjadi, Mehdi S. M. and Barron, Jonathan T. and Dosovitskiy, Alexey and Duckworth, Daniel},
  booktitle = {CVPR},
  year      = {2021}
}

@inproceedings{NeuRBF_2023_ICCV,
	author    = {Chen, Zhang and Li, Zhong and Song, Liangchen and Chen, Lele and Yu, Jingyi and Yuan, Junsong and Xu, Yi},
	title     = {{NeuRBF}: A Neural Fields Representation with Adaptive Radial Basis Functions},
	booktitle = {ICCV},
	month     = {October},
	year      = {2023},
	pages     = {4182-4194}
}

@inproceedings{levy2023seathrunerf,
  title     = {SeaThru-NeRF: Neural Radiance Fields in Scattering Media},
  author    = {Levy, Deborah and Peleg, Amit and Pearl, Naama and Rosenbaum, Dan and Akkaynak, Derya and Korman, Simon and Treibitz, Tali},
  booktitle = {CVPR},
  year      = {2023},
  doi       = {10.1109/CVPR52729.2023.00014}
}

@inproceedings{li2025watersplatting,
  title     = {WaterSplatting: Fast Underwater 3D Scene Reconstruction Using Gaussian Splatting},
  author    = {Li, Huapeng and Song, Wenxuan and Xu, Tianao and Elsig, Alexandre and Kulhanek, Jonas},
  booktitle = {3DV},
  year      = {2025}
}

@inproceedings{yang2025seasplat,
  title     = {SeaSplat: Representing Underwater Scenes with 3D Gaussian Splatting and a Physically Grounded Image Formation Model},
  author    = {Yang, Daniel and Leonard, John J. and Girdhar, Yogesh},
  booktitle = {ICRA},
  pages     = {7632--7638},
  year      = {2025},
  doi       = {10.1109/ICRA55743.2025.11128502}
}

@inproceedings{akkaynak2018revised,
  title     = {A Revised Underwater Image Formation Model},
  author    = {Akkaynak, Derya and Treibitz, Tali},
  booktitle = {CVPR},
  year      = {2018}
}

@inproceedings{seathru_cvpr_2019,
  title     = {Sea-Thru: A Method for Removing Water From Underwater Images},
  author    = {Akkaynak, Derya and Treibitz, Tali},
  booktitle = {CVPR},
  year      = {2019},
  doi       = {10.1109/CVPR.2019.00178}
}

@INPROCEEDINGS{RestorGS_cvpr_2025,
  author={Qiao, Yuanjian and Shao, Mingwen and Meng, Lingzhuang and Xu, Kai},
  booktitle={CVPR}, 
  title={{RestorGS}: Depth-aware Gaussian Splatting for Efficient 3D Scene Restoration}, 
  year={2025},
  doi={10.1109/CVPR52734.2025.01044}}

@INPROCEEDINGS{WaterNeRF_2023,
  author={Sethuraman, Advaith Venkatramanan and Ramanagopal, Manikandasriram Srinivasan and Skinner, Katherine A.},
  booktitle={OCEANS}, 
  title={{WaterNeRF}: Neural Radiance Fields for Underwater Scenes}, 
  year={2023},
  volume={},
  number={},
  pages={1-7},
  doi={10.23919/OCEANS52994.2023.10336972}}

@inproceedings{UW-GS_WACV_2025,
	title={{UW-GS}: Distractor-Aware 3D Gaussian Splatting for Enhanced Underwater Scene Reconstruction}, 
	author={Haoran Wang and Nantheera Anantrasirichai and Fan Zhang and David Bull},
	year={2025},
	booktitle={WACV},
}

@ARTICLE{yuan2025_3duir,
  author={Yuan, Jieyu and Li, Yujun and Zhang, Yuanlin and Guo, Chunle and Tang, Xiongxin and Wang, Ruixing and Li, Chongyi},
  journal={IEEE TIP}, 
  title={{3D-UIR}: {3D} Gaussian for Underwater {3D} Scene Reconstruction via Physics-Based Appearance–Medium Decoupling}, 
  year={2026},
  volume={35},
  number={},
  pages={5452-5465},
  doi={10.1109/TIP.2026.3694143}}

@inproceedings{akkaynak2017space,
  title     = {What Is the Space of Attenuation Coefficients in Underwater Computer Vision?},
  author    = {Akkaynak, Derya and Treibitz, Tali and Shlesinger, Tom and Tamir, Raz and Loya, Yossi and Iluz, David},
  booktitle = {CVPR},
  year      = {2017}
}

@article{gharbi2017deepbilateral,
  title   = {Deep Bilateral Learning for Real-Time Image Enhancement},
  author  = {Gharbi, Micha{\"e}l and Chen, Jiawen and Barron, Jonathan T. and Hasinoff, Samuel W. and Durand, Fr{\'e}do},
  journal = {ACM Transactions on Graphics},
  volume  = {36},
  number  = {4},
  year    = {2017},
  doi     = {10.1145/3072959.3073592}
}

@article{BilRF_tog_2024,
author = {Wang, Yuehao and Wang, Chaoyi and Gong, Bingchen and Xue, Tianfan},
title = {Bilateral Guided Radiance Field Processing},
year = {2024},
issue_date = {July 2024},
volume = {43},
number = {4},
issn = {0730-0301},
url = {https://doi.org/10.1145/3658148},
doi = {10.1145/3658148},
journal = {ACM TOG},
month = jul,
articleno = {148},
numpages = {13},

}

@inproceedings{dahmani2024swag,
  title     = {{SWAG}: Splatting in the Wild images with Appearance-conditioned Gaussians},
  author    = {Dahmani, Hiba and Bennehar, Moussab and Piasco, Nathan and Rold{\~a}o, Luis and Tsishkou, Dzmitry},
  booktitle = {ECCV},
  year      = {2024}
}

@InProceedings{SplatSuRe_cvpr_2026,
    author    = {Asthana, Pranav and Hanson, Alex and Tu, Allen and Goldstein, Tom and Zwicker, Matthias and Varshney, Amitabh},
    title     = {SplatSuRe: Selective Super-Resolution for Multi-view Consistent 3D Gaussian Splatting},
    booktitle = {CVPR},
    month     = {June},
    year      = {2026},
    pages     = {11840-11849},
    url       = {https://splatsure.github.io/}
}

@inproceedings{zhang2024GS-W,
  title={Gaussian in the wild: 3d gaussian splatting for unconstrained image collections},
  author={Zhang, Dongbin and Wang, Chuming and Wang, Weitao and Li, Peihao and Qin, Minghan and Wang, Haoqian},
  booktitle={ECCV},
  year={2024},
}

@article{wu2025plenodium,
    title={Plenodium: UnderWater 3D Scene Reconstruction with Plenoptic Medium Representation},
    author={Wu, Changguanng and Dong, Jiangxin and Li, Chengjian and Tang, Jinhui},
    journal={NeurIPS},
    year={2025}   
    }

@article{UIEB_TIP_2019,
  title={An underwater image enhancement benchmark dataset and beyond},
  author={Li, Chongyi and Guo, Chunle and Ren, Wenqi and Cong, Runmin and Hou, Junhui and Kwong, Sam and Tao, Dacheng},
  journal={IEEE TIP},
  volume={29},
  pages={4376--4389},
  year={2019},
  publisher={IEEE}
}

@inproceedings{kweon2026oceansplat,
  title={OceanSplat: Object-aware Gaussian Splatting with Trinocular View Consistency for Underwater Scene Reconstruction},
  author={Kweon, Minseong and Park, Jinsun},
  booktitle={AAAI},
  year={2026}
}

@inproceedings{marinestd_2025_acmmm,
author = {Liu, Shaohua and Gao, Ning and Gu, Zuoya and Dou, Hongkun and Deng, Yue and Li, Hongjue},
title = {Spatiotemporal Degradation-Aware 3D Gaussian Splatting for Realistic Underwater Scene Reconstruction},
year = {2025},
doi = {10.1145/3746027.3754888},
booktitle = {ACM MM},
pages = {141--150}
}

@article{WaterHENeRF_2024_IF,
	title = {{WaterHE-NeRF}: Water-ray matching neural radiance fields for underwater scene reconstruction},
	journal = {Inf. Fusion},
	pages = {102770},
	year = {2024},
        month = {Mar.},
	issn = {1566-2535},
	doi = {https://doi.org/10.1016/j.inffus.2024.102770},
	author = {Jingchun Zhou and Tianyu Liang and Dehuan Zhang and Siyuan Liu and Junsheng Wang and Edmond Q. Wu},
}

@InProceedings{SFM_2016_CVPR,
	author = {Schonberger, Johannes L. and Frahm, Jan-Michael},
	title = {Structure-From-Motion Revisited},
	booktitle =  {CVPR},
	month = {Jun.},
	year = {2016}
}

@inproceedings{Depth_Anything_V2,
 author = {Yang, Lihe and Kang, Bingyi and Huang, Zilong and Zhao, Zhen and Xu, Xiaogang and Feng, Jiashi and Zhao, Hengshuang},
 booktitle =  {NeurIPS},
 pages = {21875--21911},
 publisher = {Curran Associates, Inc.},
 title = {Depth Anything V2},
 volume = {37},
 year = {2024}
}

@inproceedings{nerfstudio,
	title        = {Nerfstudio: A Modular Framework for Neural Radiance Field Development},
	author       = {
	Tancik, Matthew and Weber, Ethan and Ng, Evonne and Li, Ruilong and Yi, Brent
	and Kerr, Justin and Wang, Terrance and Kristoffersen, Alexander and Austin,
	Jake and Salahi, Kamyar and Ahuja, Abhik and McAllister, David and Kanazawa,
	Angjoo
	},
	year         = 2023,
	booktitle    = {SIGGRAPH},
}

@article{ye2025gsplat,
	title={gsplat: An open-source library for Gaussian splatting},
	author={Ye, Vickie and Li, Ruilong and Kerr, Justin and Turkulainen, Matias and Yi, Brent and Pan, Zhuoyang and Seiskari, Otto and Ye, Jianbo and Hu, Jeffrey and Tancik, Matthew and Angjoo Kanazawa},
	journal={JMLR},
	volume={26},
	number={34},
	pages={1--17},
	year={2025}
}

@inproceedings{uranker_aaai_2023,
	title={Underwater ranker: Learn which is better and how to be better},
	author={Guo, Chunle and Wu, Ruiqi and Jin, Xin and Han, Linghao and Zhang, Weidong and Chai, Zhi and Li, Chongyi},
	booktitle={AAAI},
	volume={37},
	pages={702--709},
	year={2023}
}

@article{uciqe_TIP_2015,
	title={An underwater color image quality evaluation metric},
	author={Yang, Miao and Sowmya, Arcot},
	journal={IEEE TIP},
	volume={24},
	number={12},
	pages={6062--6071},
	year={2015},
	publisher={IEEE}
}

@INPROCEEDINGS{MUSIQ_iccv_2021,
  author={Ke, Junjie and Wang, Qifei and Wang, Yilin and Milanfar, Peyman and Yang, Feng},
  booktitle={ICCV}, 
  title={MUSIQ: Multi-scale Image Quality Transformer}, 
  year={2021},
  volume={},
  number={},
  pages={5128-5137},
  doi={10.1109/ICCV48922.2021.00510}}

@inproceedings{stylizednerf_cvpr_2022,
  title={Stylizednerf: consistent 3d scene stylization as stylized nerf via 2d-3d mutual learning},
  author={Huang, Yi-Hua and He, Yue and Yuan, Yu-Jie and Lai, Yu-Kun and Gao, Lin},
  booktitle={CVPR},
  pages={18342--18352},
  year={2022}
}

@InProceedings{RAFT_eccv_2020,
author="Teed, Zachary
and Deng, Jia",
title="RAFT: Recurrent All-Pairs Field Transforms for Optical Flow",
booktitle="ECCV",
year="2020",
}

@inproceedings{matspray_cvpr_2026,
  author = {Langsteiner, Philipp and Dihlmann, Jan-Niklas and Lensch, Hendrik},
  title = {{MatSpray}: Fusing {2D} Material World Knowledge on {3D} Geometry},
  booktitle = {CVPR},
  year = {2026},
  pages = {22615--22625}
}

@inproceedings{intergsedit_iccv_2025,
  author = {Wen, Minghao and Wu, Shengjie and Wang, Kangkan and Liang, Dong},
  title = {{InterGSEdit}: Interactive {3D} Gaussian Splatting Editing with {3D} Geometry-Consistent Attention Prior},
  booktitle = {ICCV},
  year = {2025},
  pages = {26136--26145}
}

@ARTICLE{LSUI_TIP_2023,
  author={Peng, Lintao and Zhu, Chunli and Bian, Liheng},
  journal={IEEE Transactions on Image Processing}, 
  title={U-Shape Transformer for Underwater Image Enhancement}, 
  year={2023},
  volume={32},
  number={},
  pages={3066-3079},
  doi={10.1109/TIP.2023.3276332}}

@article{Ucolor_TIP_2021,
  author = {Li, Chongyi and Anwar, Saeed and Hou, Junhui and Cong, Runmin and Guo, Chunle and Ren, Wenqi},
  title = {Underwater Image Enhancement via Medium Transmission-Guided Multi-Color Space Embedding},
  journal = {IEEE TIP},
  year = {2021},
  doi = {10.1109/TIP.2021.3076367}
}

@inproceedings{osmosis_eccv_2024,
  author = {Bar Nathan, Opher and Levy, Deborah and Treibitz, Tali and Rosenbaum, Dan},
  title = {{Osmosis}: {RGBD} Diffusion Prior for Underwater Image Restoration},
  booktitle = {ECCV},
  year = {2024}
}

@article{slurpp_tpami_2025,
  author = {Wu, Jiayi and Wang, Tianfu and Siddique, Md Abu Bakr and Islam, Md Jahidul and Ferm{\"u}ller, Cornelia and Aloimonos, Yiannis and Metzler, Christopher A.},
  title = {Single-Step Latent Diffusion for Underwater Image Restoration},
  journal = {IEEE TPAMI},
  year = {2026},
  volume = {48},
  number = {10},
  pages = {11324--11334},
  doi = {10.1109/TPAMI.2025.3599775}
}

@article{huber1964robust,
  author = {Huber, Peter J.},
  title = {Robust Estimation of a Location Parameter},
  journal = {The Annals of Mathematical Statistics},
  volume = {35},
  number = {1},
  pages = {73--101},
  year = {1964},
  doi = {10.1214/aoms/1177703732}
}

@inproceedings{Plenoxels_2022_CVPR,
      title={Plenoxels: Radiance Fields without Neural Networks},
      author={Sara Fridovich-Keil and Alex Yu and Matthew Tancik and Qinhong Chen and Benjamin Recht and Angjoo Kanazawa},
      year={2022},
      booktitle={CVPR},
}

@inproceedings{2dgs,
author = {Huang, Binbin and Yu, Zehao and Chen, Anpei and Geiger, Andreas and Gao, Shenghua},
title = {{2D} Gaussian Splatting for Geometrically Accurate Radiance Fields},
year = {2024},
doi = {10.1145/3641519.3657428},
booktitle = {SIGGRAPH},
}

@InProceedings{Mildenhall_2022_CVPR,
    author    = {Mildenhall, Ben and Hedman, Peter and Martin-Brualla, Ricardo and Srinivasan, Pratul P. and Barron, Jonathan T.},
    title     = {{NeRF} in the Dark: High Dynamic Range View Synthesis From Noisy Raw Images},
    booktitle = {CVPR},
    month     = {June},
    year      = {2022},
    pages     = {16190-16199}
}

@InProceedings{LE3D_2024,
	title={Lighting Every Darkness with {3DGS}: Fast Training and Real-Time Rendering for HDR View Synthesis}, 
	author={Xin Jin and Pengyi Jiao and Zheng-Peng Duan and Xingchao Yang and Chun-Le Guo and Bo Ren and Chongyi Li},
	year={2024},
	booktitle= {NeurIPS},
}

@inproceedings{cui_luminance_gs,
	  title = {Luminance-GS: Adapting 3D Gaussian Splatting to Challenging Lighting Conditions with View-Adaptive Curve Adjustment},
	  author = {Cui, Ziteng and Chu, Xuangeng and Harada, Tatsuya},
	  booktitle={CVPR},
	  year={2025}}

@InProceedings{Liu_2025_CVPR,
    author    = {Liu, Jinfeng and Kong, Lingtong and Li, Bo and Xu, Dan},
    title     = {GaussHDR: High Dynamic Range Gaussian Splatting via Learning Unified 3D and 2D Local Tone Mapping},
    booktitle = {CVPR},
    month     = {June},
    year      = {2025},
    pages     = {5991-6000}
}

@InProceedings{Mip-Splatting_2024_CVPR,
	author    = {Yu, Zehao and Chen, Anpei and Huang, Binbin and Sattler, Torsten and Geiger, Andreas},
	title     = {{Mip-Splatting}: Alias-free 3D Gaussian Splatting},
	booktitle =  {CVPR},
	month     = {June},
	year      = {2024},
	pages     = {19447-19456}
}
